\documentclass[11pt]{article}

\usepackage[preprint]{acl}

\usepackage{times}
\usepackage{latexsym}
\usepackage{hyperref} 
\usepackage{amsmath} 
\usepackage{booktabs, multirow, graphicx}
\usepackage[T1]{fontenc}

\usepackage[utf8]{inputenc}
\usepackage{pifont}
\usepackage{microtype}
\usepackage{fontawesome}

\usepackage[table]{xcolor}
\usepackage{tikz}

\definecolor{goodgreen}{RGB}{46,125,50}
\definecolor{badred}{RGB}{198,40,40}
\definecolor{okayorange}{RGB}{230,145,56}
\definecolor{neutralgray}{RGB}{120,120,120}

\newcommand{\fcircle}[1]{%
  \tikz[baseline=-0.6ex]\fill[#1] (0,0) circle (0.7ex);%
}

\newcommand{\gcirc}{\fcircle{goodgreen}}
\newcommand{\rcirc}{\fcircle{badred}}
\newcommand{\ocirc}{\fcircle{okayorange}}

\newcommand{\rothead}[1]{\rotatebox{90}{\parbox{2.6cm}{\centering\textbf{#1}}}}
\newcommand{\xmark}{\ding{55}}
\newcommand{\cmark}{\ding{51}}

\usepackage{inconsolata}
\usepackage{graphicx}
\usepackage{booktabs}  
\usepackage{url}  
\usepackage{caption}

\title{%
  \raisebox{-0.35\height}{\includegraphics[width=0.08\linewidth]{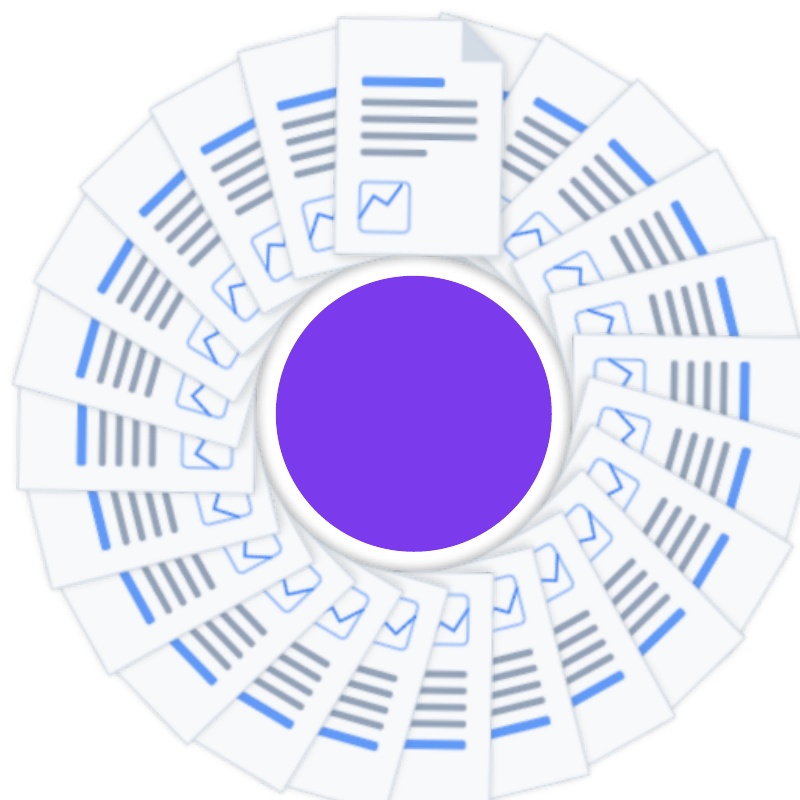}}%
  \hspace{0.3em}%
  Paper Circle: An Open-source Multi-agent Research Discovery and Analysis Framework
}

\author{
Komal Kumar$^{1}$,
Aman Chadha$^{2}$,
Salman Khan$^{1}$,
Fahad Shahbaz Khan$^{1}$,
Hisham Cholakkal$^{1}$ \\
$^{1}$ Mohamed bin Zayed University of Artificial Intelligence \\
$^{2}$ AWS Generative AI Innovation Center, Amazon Web Services \\
\small
\begin{tabular}{cc}
{\fontsize{10}{10}\selectfont\faGithub}~\textbf{GitHub:}~\href{https://github.com/MAXNORM8650/papercircle}{\texttt{\textcolor{teal}{github.com/MAXNORM8650/papercircle}}}
\end{tabular} \\[2pt]
\small
{\fontsize{10}{10}\selectfont\faGlobe}~\textbf{Website:}~\href{https://papercircle.vercel.app/}{\texttt{\textcolor{teal}{papercircle.vercel.app/}}}
}

\begin{document}


\makeatletter
\let\@oldmaketitle\@maketitle
\renewcommand{\@maketitle}{%
  \@oldmaketitle
  \vspace{0.75em}
  \begin{center}
    \includegraphics[width=\textwidth]{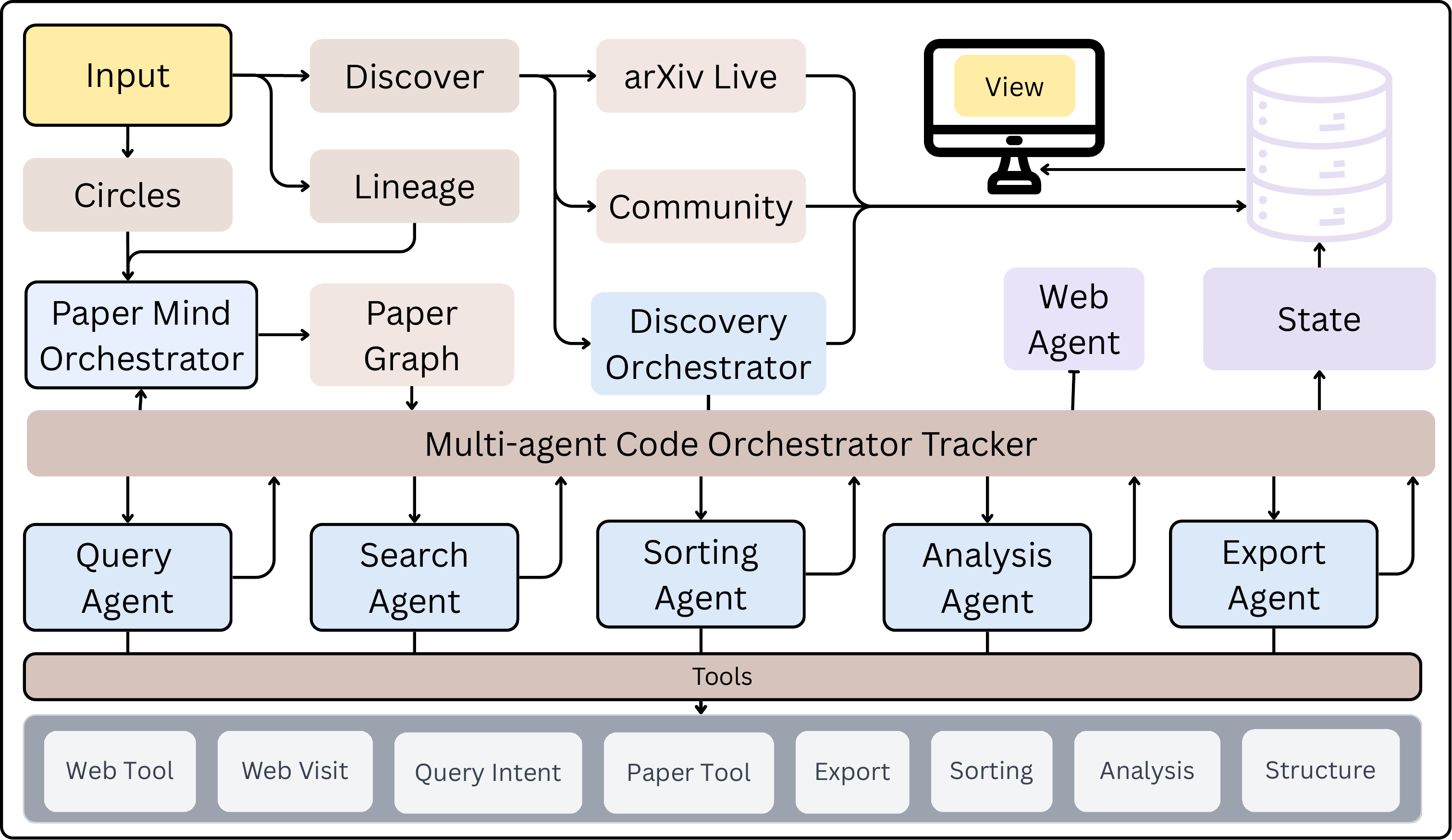}
    \captionof{figure}{
    Overview of the Paper Circle pipeline. Given a user query, Paper Circle builds a paper set from multiple sources (e.g., paper graph, community, and arXiv live) via the Paper Mind for analysis and Discovery Orchestrators for search of the paper. A multi-agent layer (query, search, sorting, analysis, export) is coordinated by the Tracker, which maintains a shared state that is persisted to a backing store and displayed to the user through interface.}
    \label{fig:main:pca}
  \end{center}
  \vspace{0.75em}
}
\makeatother
\maketitle
\begin{abstract}
The rapid growth of scientific literature has made it increasingly difficult for researchers to efficiently discover, evaluate, and synthesize relevant work. Recent advances in multi-agent large language models (LLMs) have demonstrated strong potential for understanding user intent and are being trained to utilize various tools.
In this paper, we introduce Paper Circle, a multi-agent research discovery and analysis system designed to reduce the effort required to find, assess, organize, and understand academic literature. The system comprises two complementary pipelines: (1) a Discovery Pipeline that integrates offline and online retrieval from multiple sources, multi-criteria scoring, diversity-aware ranking, and structured outputs; and (2) an Analysis Pipeline that transforms individual papers into structured knowledge graphs with typed nodes (e.g., concepts, methods, experiments, and figures) and edges, enabling graph-aware question answering and coverage verification. Both pipelines are implemented within a coder LLM–based multi-agent orchestration framework and produce fully reproducible, synchronized outputs (JSON, CSV, BibTeX, Markdown, and HTML) at each agent step. This paper describes the system architecture, agent roles, retrieval and scoring methods, knowledge graph schema, and evaluation interfaces that together form the Paper Circle research workflow. We benchmark Paper Circle on both paper retrieval and paper review generation, reporting hit rate, MRR, and Recall@K. Results show consistent improvements with stronger agent models. We have publicly released the \href{https://papercircle.vercel.app/}{website} and \href{https://github.com/MAXNORM8650/papercircle}{code}.
\end{abstract}

\section{Introduction}

The pace of scientific publication has accelerated exponentially, creating a significant burden on researchers attempting to stay abreast of new developments \cite{reddy2025_30, pramanick2023_39}. Traditional search engines and recommendation systems often struggle to provide the depth and context required for rigorous literature reviews, leading to fragmented discovery workflows. Recently, the advent of Large Language Models (LLMs) has catalyzed a shift towards "AI Scientists", autonomous multi-agent systems (MAS) capable of generating hypotheses, conducting experiments, and even writing papers \cite{chen2025_10, naumov2025_51}. While these systems demonstrate the potential of agentic workflows, there remains a critical gap between fully autonomous simulations and the practical, collaborative needs of human research communities.

Paper Circle addresses (as shown in the Figure \ref{fig:main:pca}) this gap by introducing a comprehensive \textit{Multi-Agent Research Platform} that supports the entire lifecycle of literature engagement: from discovery and analysis to critique and synthesis.
In the Table~\ref{tab:paper_circle_comparison}, we compared to existing multi-agent architectures for scientific literature tasks. Paper Circle offers a unique combination of capabilities that no existing system jointly provides. Specifically, it is designed to reduce the effort required to find, assess, organize, and understand academic literature.

Unlike purely autonomous systems that aim to replace the researcher, Paper Circle is designed as a collaborative workbench that augments human intelligence through three integrated subsystems:
\begin{table}[t]
\centering
\caption{Comparison of Paper Circle against prior literature systems. Green indicates supported, orange indicates partial support, and red indicates unsupported.}
\label{tab:paper_circle_comparison}
\small
\renewcommand{\arraystretch}{1}
\setlength{\tabcolsep}{6pt}

\resizebox{0.5\textwidth}{!}{%
\begin{tabular}{lcccccccc}
\toprule
\textbf{System} &
\rothead{Multi-agent Orchestration} &
\rothead{Multi-source Discovery} &
\rothead{Typed Paper KG} &
\rothead{Node/Edge Provenance} &
\rothead{Coverage Verification} &
\rothead{Graph-aware QA} &
\rothead{Deterministic Runs} &
\rothead{Structured Exports (bib,csv,md etc)} \\
\midrule
\textbf{Paper Circle}     & \gcirc & \gcirc & \gcirc & \gcirc & \gcirc & \gcirc & \gcirc & \gcirc \\
\textbf{PaperQA~\cite{lala2023paperqa}}          & \ocirc & \rcirc & \rcirc & \ocirc & \rcirc & \rcirc & \ocirc & \ocirc \\
\textbf{PaperQA2}~\cite{lala2023paperqa}         & \ocirc & \rcirc & \rcirc & \ocirc & \rcirc & \rcirc & \ocirc & \ocirc \\
\textbf{STORM}~\cite{shao-etal-2024-assisting}            & \gcirc & \ocirc & \rcirc & \ocirc & \rcirc & \rcirc & \ocirc & \ocirc \\
\textbf{SciSage}~\cite{shi2025scisage}         & \gcirc & \ocirc & \rcirc & \ocirc & \rcirc & \rcirc & \ocirc & \ocirc \\
\textbf{Con.Papers}~\href{https://www.connectedpapers.com/}{connectedpapers.com}& \rcirc & \gcirc & \rcirc & \rcirc & \rcirc & \rcirc & \gcirc & \rcirc \\
\textbf{alphaXiv}~\href{https://www.alphaxiv.org/}{alphaxiv.org}        & \rcirc & \ocirc & \rcirc & \rcirc & \rcirc & \ocirc & \ocirc & \rcirc \\
\bottomrule
\end{tabular}%

}

\vspace{0.4em}
\noindent\small
\gcirc~Favorable \qquad \ocirc~Partial \qquad \rcirc~Unfavorable
\end{table}

\begin{enumerate}
    \item \textbf{Discovery Pipeline}: A multi-agent retrieval system that goes beyond simple keyword matching. It employs a multi-dimensional scoring framework to surface high-value research. Crucially, this pipeline is deterministic and produces structured artifacts (JSON, linear logs) at every step.
    \item \textbf{Paper Mind Graph}: To facilitate deep understanding, Paper Circle constructs a dynamic Knowledge Graph from retrieved literature. This "Paper Mind" enables researchers to query the collective intelligence of a reading list, identifying latent connections between disparate works and supporting complex Question-Answering workflows that are grounded in specific citation sub-graphs.
    \item \textbf{Review Agents}: This platform features a team of specialized review agents that generate detailed critiques and scores, consistently highlighting strengths and weaknesses to guide human reading priorities \cite{naumov2025_51}.
\end{enumerate}

By integrating these capabilities into a shared "Reading Circle" environment, Paper Circle transforms literature review from a solitary task into a community-driven, AI-augmented operation.

\section{Related Work}

\subsection{Autonomous Scientific Discovery}
The emerging field of AI-Scientists aim to automate the entire research lifecycle. Systems like DORA AI agent \cite{naumov2025_51} and EvoResearch \cite{gajjar2025_61} demonstrate end-to-end capabilities, from hypothesis generation to report writing. Similarly, O-Researcher \cite{unknown2026_19}, MARS \cite{unknown2026_20}, and AlphaResearch \cite{unknown2026_17} treat research as a multi-step optimization problem, often using reinforcement learning to refine discovery strategies. Specialized agents have also been proposed for causal discovery, such as CausalSteward \cite{unknown2026_7} and other multi-agent frameworks \cite{le2025_0}. While these systems push the boundaries of autonomy, Paper Circle prioritizes \textit{curation and reproducibility} over full automation. Instead of replacing the researcher, Paper Circle acts as a force multiplier for human teams, ensuring that the discovery process remains transparent and verifiable.

\subsection{MAS in Specialized Domains}
MAS have shown remarkable success in specific scientific verticals. In chemistry and materials science, frameworks like ChemThinker \cite{ju2025_16}, MOOSE-Chem \cite{yang2025_25}, and ChemBOMAS \cite{unknown2026_26} leverage LLMs to discover new molecules and optimize experiments \cite{kumbhar2025_37}. In biology and healthcare, agents facilitate single-cell analysis (CellAgent \cite{unknown2026_2}), phenotype discovery (PhenoGraph \cite{niyakan2025_52}), and clinical data analysis \cite{spieser2025_53}. Other applications range from drug discovery \cite{fehlis2025_56} and psychiatry diagnosis \cite{xiao2025_34} to financial forecasting, where systems like ASTRAFIN \cite{singh2025_57} and other stock analysis agents \cite{chandrashekar2025_47, wawer2025_50} predict market trends. Paper Circle complements these domain-specific tools by providing a \textit{general-purpose} discovery pipeline that can be adapted to any discipline, serving as the foundational layer for literature review and knowledge management.

\subsection{Community Simulation and Collaboration}
A distinct line of research focuses on simulating or facilitating the social aspects of science. ResearchTown \cite{yu2025_1, yu2025_3} models the research community using agents to understand how ideas propagate. Other works explore collaborative dynamics through automated negotiation (NegoLog \cite{doru2024_21}, NEGOTIATOR \cite{keskin2024_31}) and cohesive dialogue generation \cite{chu2024_27}. Frameworks like PiFlow \cite{unknown2026_6}, REDEREF \cite{unknown2026_22}, and blackboard systems \cite{unknown2026_5} propose mechanisms for agent collaboration in information discovery. Paper Circle distinguishes itself by moving beyond simulation; it provides a real-world platform for \textit{human-AI collaboration}. It does not just model how researchers might interact, but actively facilitates those interactions through shared reading lists, discussion threads, and collaborative ranking.


\section{Methodology}

\subsection{Background}

Multi-Agent Systems (MAS) represent a paradigm where autonomous entities interact to solve complex problems distributedly. In the context of scientific discovery, MAS allows for the decomposition of intricate research tasks,such as literature search, reading, and reasoning,into manageable sub-routines handled by specialized agents \cite{wooldridge2002}. Unlike monolithic LLM approaches, agentic workflows can maintain distinct personas (e.g., "The Skeptic", "The Creative") and leverage external tools, reducing hallucination and improving reasoning depth through inter-agent dialogue \cite{reddy2025_30}.

The baseline for our orchestration layer is the \texttt{smolagents} \citep{smolagents} library. The pipeline uses a \texttt{CodeAgent (CoA)} as the central orchestrator, which can attend parallel agent calls and toll calls and multiple \texttt{ToolCallingAgent (ToCA)} instances, each attached to specific capabilities (e.g., arXiv retrival, PDF parsing). The baseline responsibilities include (i) tool invocation, (ii) multi-step planning via the orchestrator, and (iii) delegation to specialized agents. PaperCircle extends this foundation by adding structured outputs, offline search capabilities, and rigorous evaluation metrics. We preserve the baseline tool interface, where each tool receives explicit parameters and returns a formatted string response, allowing the orchestrator to chain steps while maintaining high readability and traceability.

\subsection{System Architecture}
Figure~\ref{fig:main:pca} illustrates the overall architecture of Paper Circle. The system consists of two complementary multi-agent pipelines: the \textit{Discovery Pipeline} for finding relevant papers, and the \textit{Analysis Pipeline} for deep understanding of individual papers.

\subsection{Paper Discovery Agent Design}
The main diagram of the discovery subsystem is shown in Figure \ref{fig:discovery}, which is composed of multiple agents, each bound to a small, explicit tool interface. It is inspired by the TTD-DR~\citep{han2025deep} for iteratively updating the updated version at each agentic step.
The core agents are:
\begin{figure}[t]
    \centering
    \includegraphics[width=\linewidth]{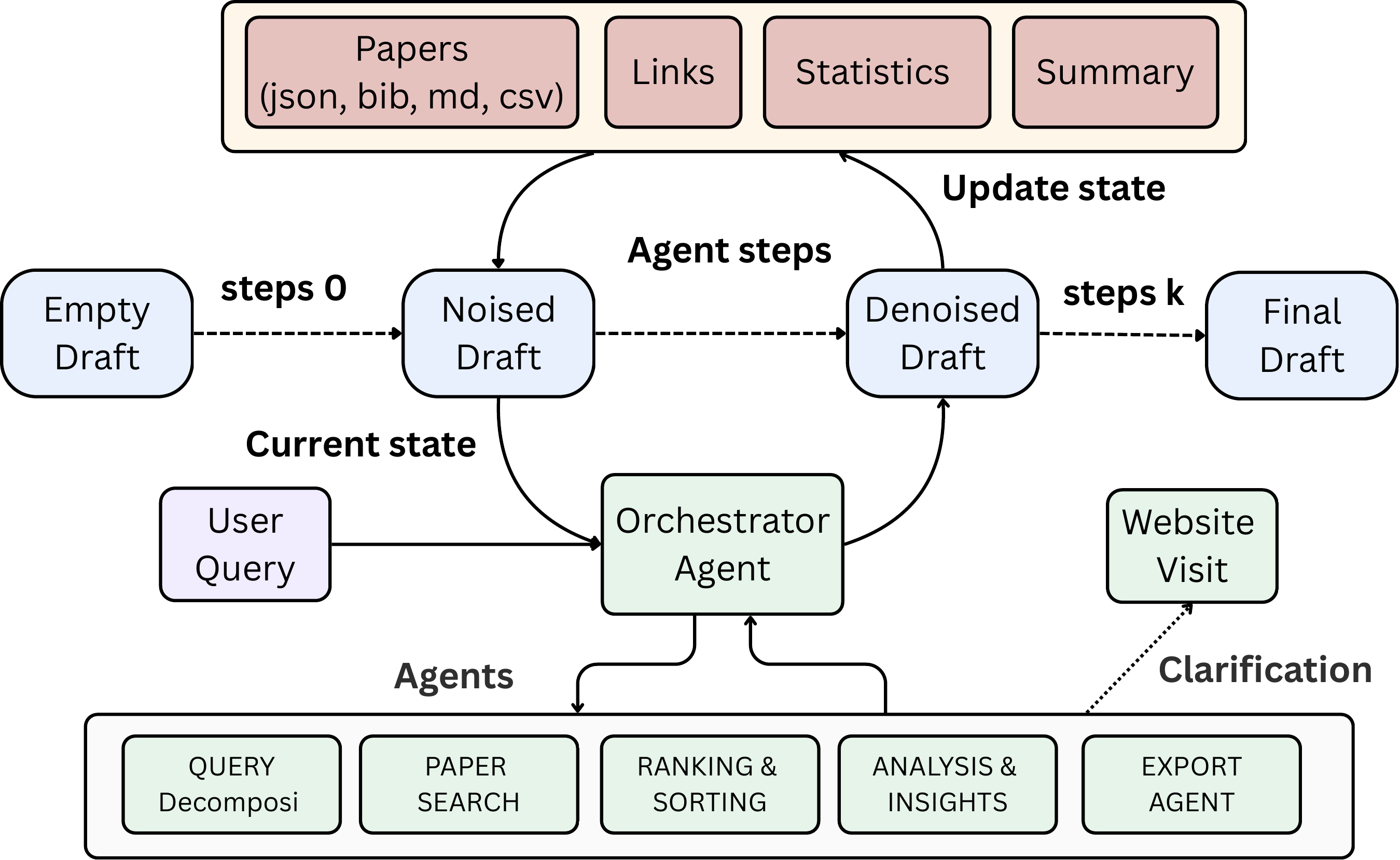}
    \caption{The main iterative diagram for the paper discovery framework. The system maintains an explicit, evolving discovery state (papers, links, statistics, and summaries) that is iteratively updated through agentic steps. Starting from an empty draft, the orchestrator agent alternates between noising and denoising operations over multiple steps, progressively refining the draft into a final result. 
    When necessary, a web search agent is invoked for clarification or recent information.
}
    \label{fig:discovery}
\end{figure}

\noindent\textbf{Intent Classification Agent.} Parses user text into search mode (offline, online, or both), conference filters, year range, and ranking preferences. Most importantly, it uses a web agent in the pipeline for any unclear queries or recent knowledge.\\
\textbf{Paper Search Agent.} Executes offline or online retrieval based on intent, merges results, performs deduplication, and updates state and outputs.\\
\textbf{Sorting Agent.} Reorders papers using recency, citations, similarity, novelty, BM25~\cite{chen2023bm25}, or combined weights; or applies a cross-encoder reranker~\cite{wang2020minilm}.\\
\textbf{Analysis Agent.} Computes aggregate statistics and insights, including source distribution, year trends, and top authors.\\
\textbf{Export Agent.} Produces synchronized exports and provides a consistent interface for downstreaming.\\
\textbf{Web Search Agent.} Provides auxiliary access to web search tools when online lookups are required.\\


\subsection{Paper Analysis Agent}\label{sec:pca}

While the discovery pipeline addresses the challenge of finding relevant papers, researchers also need to understand and synthesize the content of individual papers deeply \cite{korat2025_58}. Paper Circle addresses this with a complementary \textit{Paper Analysis Agent} that transforms research papers into structured, queryable knowledge graphs with full traceability to the original text.
\begin{figure}
    \centering
    \includegraphics[width=\linewidth]{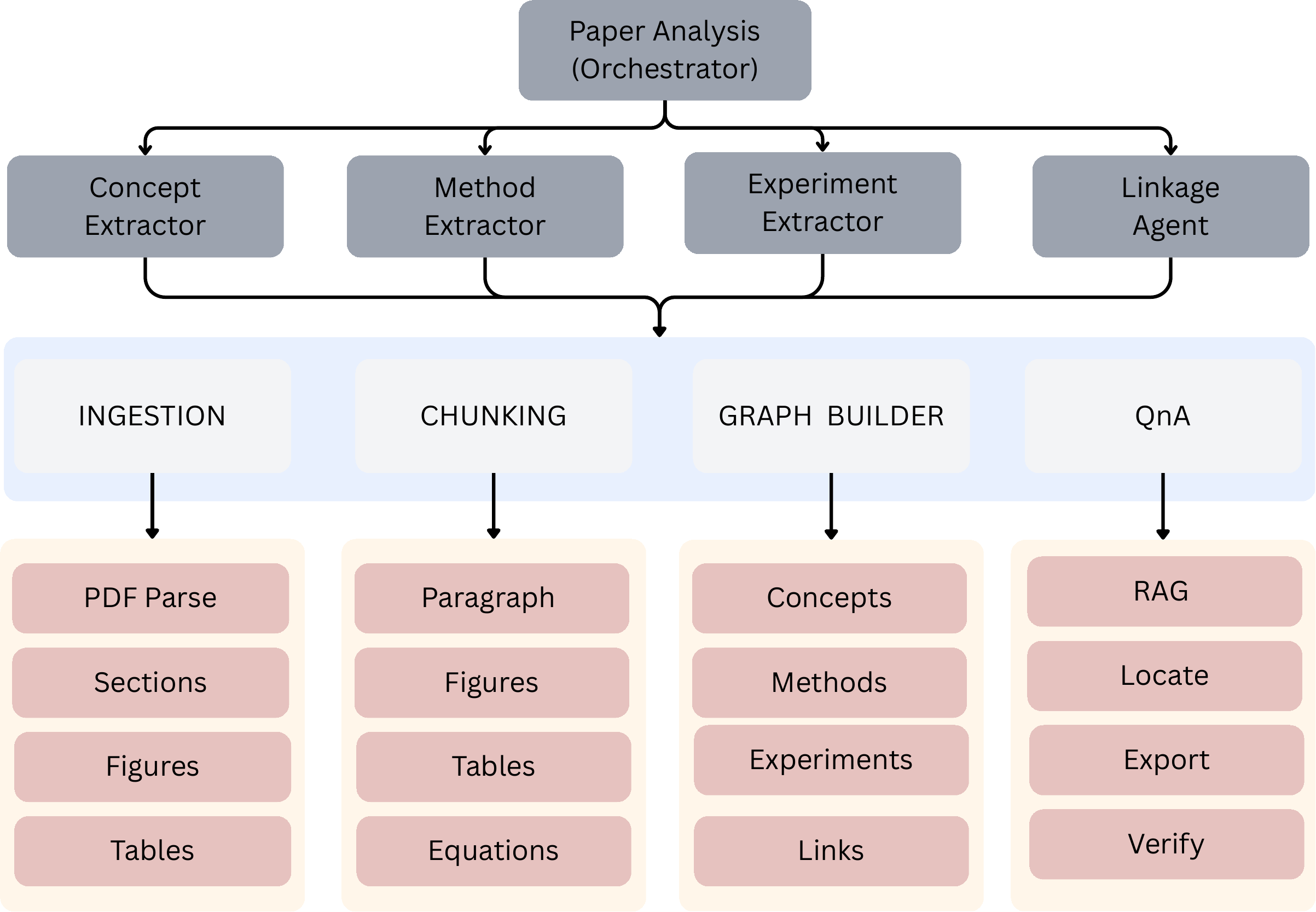}
    \caption{A paper analysis orchestrator agents for concepts, methods, experiments, and cross-entity linkages. The pipeline consists of four main stages: ingestion, which parses PDFs into structured elements (sections, figures, tables, equations); semantic chunking, which produces structure-aware text units; graph construction, which builds a typed knowledge graph of concepts, methods, experiments, and their relations with full traceability to source text; and a Q\&A layer that enables graph-aware retrieval, verification, and export. }
    \label{fig:Paper:analysis}
\end{figure}
The Paper Analysis Agent operates as a multi-stage pipeline with four specialized components as shown in the figure: (1) Ingestion Layer, (2) Graph Builder, (3) Q\&A System, and Verification Layer.


\paragraph{PDF Ingestion and Chunking.} The ingestion pipeline uses PyMuPDF for robust PDF parsing~\cite{adhikari2024comparative}. The \texttt{PDFParser} class extracts: \noindent\textbf{Metadata}: Title, authors, abstract, arXiv ID, venue, and page count.
\noindent\textbf{Sections}: Hierarchical section structure with parent-child relationships, identified via numbering patterns (e.g., ``1.2 Background'').
\noindent\textbf{Figures and Tables}: Caption text, page locations, and nearby context for linkage.
\noindent\textbf{Equations}: Numbered equations with surrounding context.

Unlike token-based chunking, the \texttt{SemanticChunker}~\cite{qu2025semantic} creates chunks aligned with document structure. Paragraphs within sections are grouped up to a configurable limit (default 1500 characters), while figures, tables, and equations are preserved as distinct chunks with their captions and context. 

\paragraph{Knowledge Graph Schema.}
The mind graph follows a typed schema with nodes~\cite{zhang2025schema} for papers, sections, concepts, methods, experiments, datasets, and visual elements (figures, tables, equations), and edges encoding structural and semantic relations (e.g., hierarchy, definition, proposal, usage, evaluation, illustration, dependency). All nodes and edges carry provenance metadata—including source chunk IDs, page numbers, verification status, confidence scores, and timestamps—ensuring full traceability to the original PDF.
\subsection{Multi-Agent Extraction}
The \texttt{GraphBuilder}~\citep{zhu2024llms} orchestrates four specialized \texttt{CoA}-based extractors. The \emph{Concept Extractor} identifies and classifies key concepts by type and importance; the \emph{Method Extractor} extracts algorithms and techniques from method sections; the \emph{Experiment Extractor} recovers experimental setups, datasets, metrics, and results; and the \emph{Linkage Agent} connects figures and tables to the concepts or methods they illustrate. Extraction proceeds in staged phases—concepts, methods, experiments, visual linkage, and inter-concept relations—each incrementally updating the shared \texttt{MindGraph}.

\paragraph{Graph-Aware Q\&A.}
The Q\&A module combines vector retrieval with graph traversal. An \texttt{EmbeddingStore} indexes text chunks and node descriptions, while the \texttt{GraphRetriever} retrieves top-$k$ relevant nodes and chunks and expands context via 1-hop neighbors. The \texttt{PaperQA} agent generates answers grounded in retrieved text, graph relations, and linked figures or tables, and returns supporting evidence with confidence estimates. A \texttt{locate} function enables precise localization of concepts, figures, or tables by page and context.

\paragraph{Coverage Verification.}
To prevent silent omissions, a \texttt{CoverageChecker} evaluates figure, table, section, and equation coverage, producing an overall coverage score and identifying unlinked or missing elements with actionable diagnostics. This provides a lightweight quality assurance step prior to downstream use.



\begin{figure}
    \centering
    \includegraphics[width=\linewidth]{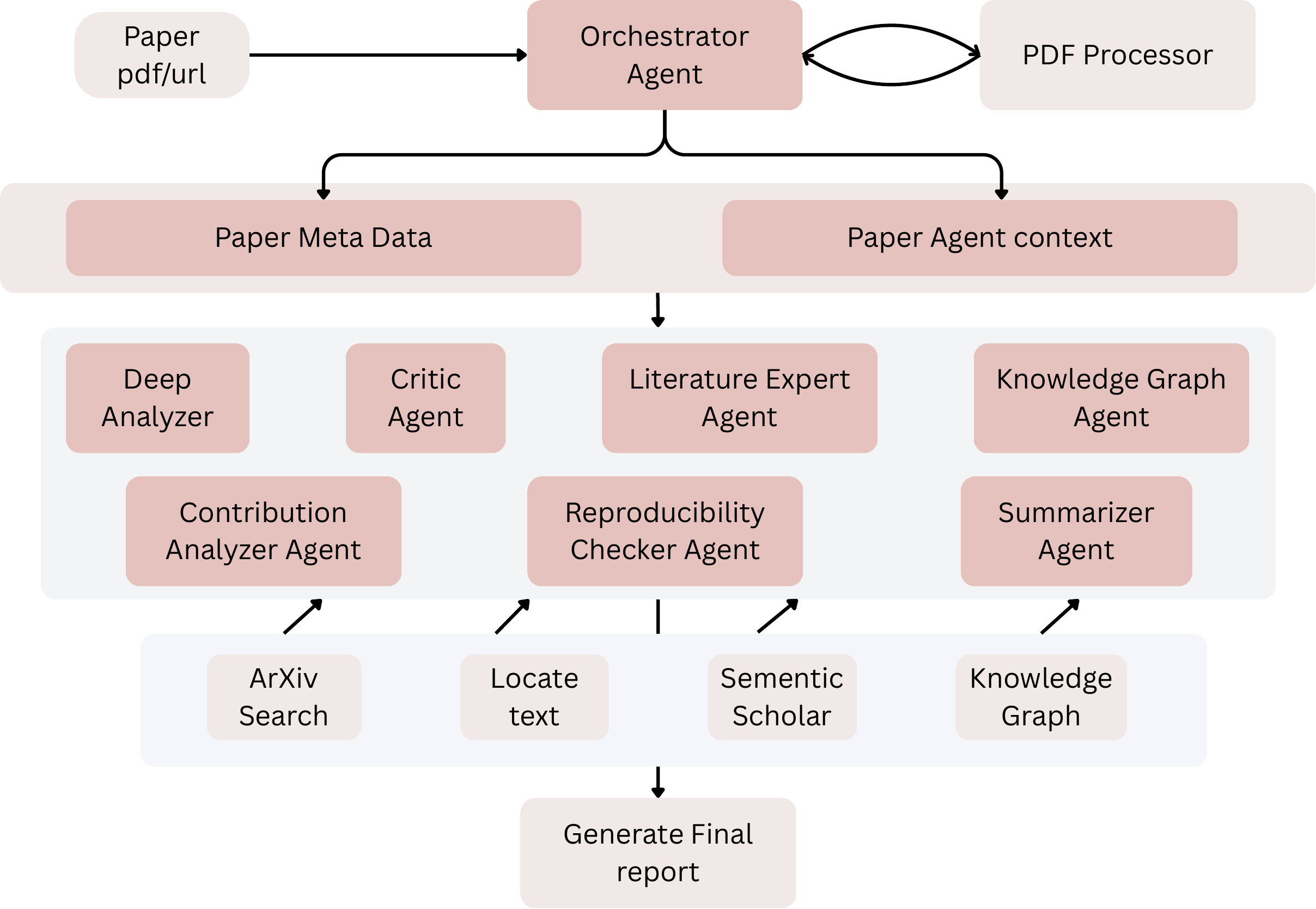}
    \caption{Multi-agent paper analysis and review architecture.
Given a paper specified by a PDF or URL, an orchestrator agent coordinates PDF processing and maintains shared paper metadata and agent context. Specialized agents operate in parallel to perform deep technical analysis, contribution extraction, critical review, literature linking, reproducibility checking, summarization, and knowledge graph construction. External tools such as arXiv search, Semantic Scholar, and targeted text localization are invoked as needed. The orchestrator aggregates agent outputs into a unified, structured final report, enabling comprehensive, reviewer-style analysis with modular extensibility.}
    \label{fig:review}
\end{figure}

\subsection{Research Review Framework}
In Sec. \ref{sec:pca}, we describe the paper analysis of agentic capabilities, which we further extend for automated peer-review-style assessment. Unlike AgentReview~\cite{jin2024agentreview,d2024marg}, we follow the paper analysis perspective, which not only provides the review but also builds a strong graph between the concepts.
\paragraph{Architecture.}
The system is built upon a multi-agent orchestration framework (Figure~\ref{fig:review}) that coordinates the execution of seven specialized roles. Each agent is instantiated as a \texttt{ToCA} or \texttt{CoA}~\cite{smolagents}.

\paragraph{Deep Analyzer.} Focuses on the technical core of the paper. It breaks down the mathematical foundations, identifies specific methodology components, and extracts primary experimental findings.
\paragraph{Critic.} Emulates a senior conference reviewer (e.g., NeurIPS, ICML). It provides a rigorous assessment of strengths and weaknesses, generates author-facing questions, and assigns scores for novelty, clarity, and significance.
\paragraph{Literature Expert.} \textls[-12]{Interfaces with external academic databases including Semantic Scholar and arXiv. It maps the paper's position within the existing research landscape and verifies citation accuracy.}
\paragraph{Contribution Analyzer.} Separates explicit author claims from verified technical contributions, identifying potential overclaiming or missing baseline comparisons.
\paragraph{Reproducibility Checker:} Quantifies the transparency of the research by assessing the availability of source code, hyperparameter specifications, dataset accessibility, and compute requirement disclosures.
\paragraph{Summarizer.} Generates multi-fidelity summaries across different abstraction levels, ranging from concise executive summaries to deep technical precis.

\paragraph{Orchestration and Pipeline Execution}
The \texttt{Multi Agent Orchestrator} manages the lifecycle of these agents through a multi-stage pipeline. The system supports parallel execution using a \texttt{ThreadPoolExecutor}.



\section{Experiments}
\subsection{Experimental setup}
All the experiments are done with open-source model with $4\times40$ GB Nvidia GPUs. We used the Ollama\footnote{\url{https://ollama.com/}} platform with the fastllm library~\citep{gong2025past}. 
\paragraph{Database Curation.}
We curated a diverse corpus, as shown in Table~\ref{tab:dataset_stats_horizontal} of research papers from leading CS and ML conferences, primarily sourced from OpenReview\footnote{\url{https://openreview.net/}} and augmented with metadata and peer-review information.
\begin{table*}[t]
    \centering
    \resizebox{\textwidth}{!}{%
    \begin{tabular}{l c c c c c c c c c c c}
        \toprule
        \textbf{Conference} 
        & ICLR 
        & NeurIPS 
        & ICML 
        & CVPR 
        & IROS 
        & ICRA 
        & AAAI 
        & ACL 
        & ICCV 
        & EMNLP 
        & Other  \\
        \midrule
        \textbf{Count}
        & 12 
        & 39 
        & 13 
        & 13 
        & 25 
        & 25 
        & 5 
        & 5 
        & 7 
        & 4 
        & 144 \\
        \bottomrule
    \end{tabular}
    }
    \caption{The Database corpus across major conferences. The ``Other'' category includes venues such as AISTATS, RSS, SIGGRAPH, and WACV. \textbf{Count} indicates the number of the most recent conference venue included.}
    \label{tab:dataset_stats_horizontal}
\end{table*}

\paragraph{Evaluation.} Paper Circle provides built-in evaluation metrics. When a ground-truth paper title or identifier is provided, the system computes Mean Reciprocal Rank (MRR), Recall@K, Precision@K, and hit rates. These metrics are computed per step and stored in the JSON file for longitudinal tracking. For batch evaluation, a parallel benchmarking utility executes multiple queries concurrently and aggregates mean metrics and timing statistics. This supports lightweight comparisons between search configurations (offline vs. online, BM25~\cite{chen2023bm25} vs. semantic (all-MiniLM-L6-v2~\cite{wang2020minilm}), with or without Qwen3-Reranker-0.6B~\citep{qwen3embedding}) without requiring external tooling.
\paragraph{Baseline Agent.} This framework is developed using the Smolagent multi-agent tool, calling the (ToCA) agent and the code agent (CoA), with tools utilized being manually developed.  
\paragraph{Architecture.}
We evaluate multiple retrieval baselines: \texttt{bm25}, \texttt{bm25+reranker} (BM25~\cite{chen2023bm25}\& cross-encoder~\cite{qwen3embedding}), \texttt{reranker}~\cite{qwen3embedding}, \texttt{semantic}~\cite{wang2020minilm}, and \texttt{hybrid} (BM25 combined with semantic retrieval). We also compare pipeline structures with different agent compositions: \texttt{full} includes all five agents (intent, search, sort, analysis, export), \texttt{minimal} uses only the search agent, \texttt{search\_sort} uses search and sort, \texttt{search\_analysis} uses search and analysis, and \texttt{no\_intent} is a full pipeline with no intent.

\subsection{Results}
\label{sec:results}
\paragraph{Natural Text-based retrieval.}
We evaluate our multi-agent paper retrieval system across multiple LLMs and retrieval baselines. We did two query type experiments, one a research assistant-based natural queries generated by running gpt-oss-20B models (called RAbench), and randomly sampling one paper record from the database, extracting a concise ``topic" phrase from its title, keywords or abstract, then picking a natural-language template and optional prefix to turn that topic into a realistic search query. We also randomly chose a scope (conference/year/range/none) to add corresponding text to the query and to emit matching structured filters. This query we referred to as SemanticBench.

All experiments were conducted on a 50 query benchmark, measuring the success rate, the hit rate, the mean reciprocal rank (MRR), and the recall.

\paragraph{Model Comparison.}
\label{subsec:model_comparison}

Table~\ref{tab:combined_results} presents comprehensive evaluation results comparing agent-based models with retrieval baselines. The results reveal a clear performance hierarchy across methods and scales. Two agent models achieve the highest retrieval effectiveness with an 80\% hit rate, qwen3-coder-30b-Q3KM (quantized) and \texttt{qwen3-coder:30b}—with qwen3-coder-30b-Q3KM also delivering the best ranking quality (MRR = 0.627) while requiring less memory for smolagent multi-step reasoning. These top-performing models are also the fastest, taking approx. 21--22 seconds per query, indicating no latency penalty for improved accuracy. The BM25 baseline remains highly competitive (78\% HR), outperforming most agent-based approaches and highlighting the continued strength of lexical matching in academic retrieval. 
Finally, RA-Bench results show higher performance than SemanticBench, suggesting that LLM-perturbed queries may be easier for multi-agent retrieval, though this requires further investigation.

\begin{table*}[htbp]
\centering
\caption{Combined benchmark results for agent-based models and retrieval baselines. Best results are shown in \textbf{bold}. All the results are calculated using semantic benchmarks. Only the last (\textcolor{blue}{blue}) is evaluated on 500 RAbench queries, which shows syntetically written query is easier to retrieve compared to the random template following.}
\label{tab:combined_results}
\resizebox{\textwidth}{!}{%
\begin{tabular}{@{}lccccccccccc@{}}
\toprule
\textbf{Model/Method} & \textbf{Type} & \textbf{Success} & \textbf{Hit Rate} & \textbf{MRR} & \textbf{R@1} & \textbf{R@5} & \textbf{R@10} & \textbf{R@20} & \textbf{R@50} & \textbf{Time (s)} & \textbf{Steps} \\
\midrule
\href{https://ollama.com/qooba/qwen3-coder-30b-a3b-instruct:q3_k_m/blobs/30c83da425db}{Qwen3C-30B-Inst-Q3\_K\_M} & Agent & 100\% & \textbf{0.80} & \textbf{0.627} & \textbf{0.58} & \textbf{0.66} & \textbf{0.74} & \textbf{0.78} & \textbf{0.80} & 22.2 & 1.42 \\
qwen3-coder:30b~\cite{qwen3technicalreport} & Agent & 100\% & \textbf{0.80} & 0.518 & 0.46 & 0.52 & 0.72 & 0.76 & \textbf{0.80} & \textbf{21.1} & 1.34 \\
\midrule
BM25~\cite{chen2023bm25} & Baseline & 100\% & 0.78 & 0.541 & 0.48 & 0.60 & 0.66 & 0.78 & 0.78 & -- & -- \\
\midrule
\href{https://www.ollama.com/sparksammy/microcoder-nonqwen3:dsr1/blobs/6ee91b1f7820}{microcoder-deepseekr1-14.8} & Agent & 52\% & 0.73 & 0.453 & 0.38 & 0.46 & 0.65 & 0.69 & 0.73 & 107.4 & 4.15 \\
deepseek-coder-v3:16b~\cite{zhu2024deepseek} & Agent & 100\% & 0.66 & 0.396 & 0.32 & 0.46 & 0.52 & 0.60 & 0.66 & 47.9 & 1.54 \\
qwen2.5-coder:3b~\citep{hui2024qwen2} & Agent & 94\% & 0.60 & 0.366 & 0.28 & 0.45 & 0.53 & 0.55 & 0.57 & 210.4 & 1.51 \\
qwen2.5-coder:14b~\citep{hui2024qwen2} & Agent & 82\% & 0.56 & 0.461 & 0.41 & 0.51 & 0.51 & 0.56 & 0.56 & 73.4 & 1.05 \\
Semantic~\citep{wang2020minilm}& Baseline & 100\% & 0.54 & 0.279 & 0.22 & 0.32 & 0.38 & 0.52 & 0.54 & -- & -- \\
Simple (bag-of-words) & Baseline & 100\% & 0.54 & 0.279 & 0.22 & 0.32 & 0.38 & 0.52 & 0.54 & -- & -- \\
qwen2.5-coder:7b~\citep{hui2024qwen2} & Agent & 100\% & 0.54 & 0.311 & 0.26 & 0.36 & 0.40 & 0.52 & 0.54 & 59.3 & 0.84 \\
\href{https://ollama.com/qooba/qwen3-coder-30b-a3b-instruct:q3_k_m/blobs/30c83da425db}{Qwen3C-30B-Inst-Q3\_K\_M} & Agent & 100\% & 0.42 & 0.348 & 0.32 & 0.38 & 0.38 & 0.40 & 0.42 & 22.7 & 1.40 \\
deepseek-coder:33b~\cite{zhu2024deepseek} & Agent & 100\% & 0.12 & 0.087 & 0.08 & 0.08 & 0.12 & 0.12 & 0.12 & 180.4 & 0.14 \\
\href{https://ollama.com/relational/orlex:latest}{qwen3vl-4b-orlex} & Agent & 12\% & 0.08 & 0.080 & 0.08 & 0.08 & 0.08 & 0.08 & 0.08 & 37.9 & 0.14 \\
granite-code:34b~\citep{mishra2024granite} & Agent & 100\% & 0.02 & 0.010 & 0.00 & 0.02 & 0.02 & 0.02 & 0.02 & 111.3 & 0.04 \\
Hybrid (BM25+sementic) & Baseline & 100\% & 0.02 & 0.001 & 0.00 & 0.00 & 0.00 & 0.00 & 0.02 & -- & -- \\
qwen2.5-coder:1.5b~\citep{hui2024qwen2} & Agent & 100\% & 0.00 & 0.000 & 0.00 & 0.00 & 0.00 & 0.00 & 0.00 & 63.7 & 0.00 \\
\href{https://ollama.com/sparksammy/microcoder-nonqwen3:gpt-oss}{microcoder-oss-20b} & Agent & 54\% & 0.00 & 0.000 & 0.00 & 0.00 & 0.00 & 0.00 & 0.00 & 47.6 & 0.00 \\
\midrule
\textcolor{blue}{Qwen3-Coder-30B-A3B-Inst-Q3\_K\_M} & Agent & 100\% & 0.98 & 0.882 & 0.83 & 0.93 & 0.95 & 0.96 & 0.97 & 21.53 & 1.36 \\
\bottomrule
\end{tabular}%
}
\end{table*}

\paragraph{Paper analysis visualization.}
In the Figure~\ref{fig:Paper:analysis}, we provide various output visualizations, including concept built graph (A), concept definition chart (B), interactive Q\&A with precise information (C), markdown analysis output (D), and finally flow chart connecting the concepts of blocks (E). All of this analysis togather provides the complete understanding of the paper.
\begin{figure*}[ht]
    \centering
    \includegraphics[width=\linewidth]{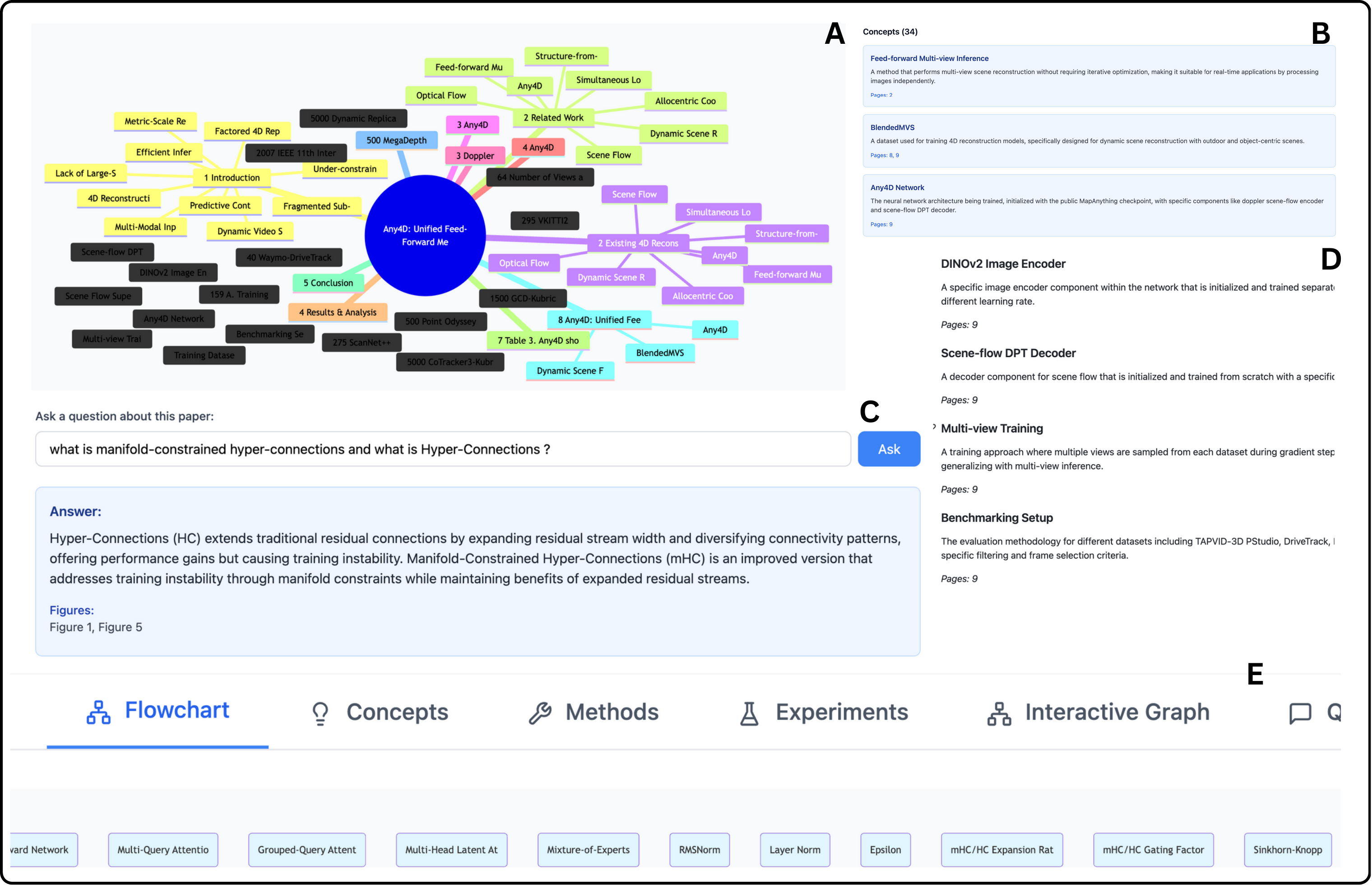}
    \caption{The main outputs of the analysis agent for a representative paper.
(A) Interactive concept graph constructed from the paper, where nodes correspond to extracted concepts and edges denote semantic relationships. (B) Automatically generated concept explanations, each linked to the originating paper sections and pages. (C) Graph-aware question answering interface, providing answers grounded in extracted content along with supporting figures and references. (D) Structured Markdown exports summarizing all extracted concepts and methods for downstream use. (E) Flowchart view illustrating the high-level organization and relationships among concepts, methods, and experimental components of the paper.}
    \label{fig:analysis_paper}
\end{figure*}

\begin{figure*}[h]
    \centering
    \includegraphics[width=\linewidth]{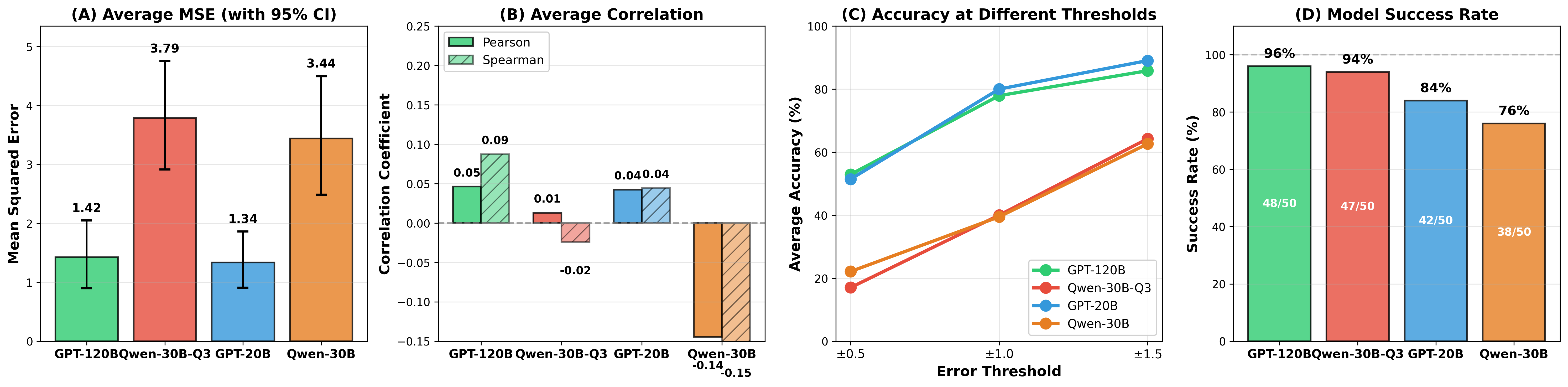}
    \caption{Paper review results analysis. This study was conducted on 50 randomly selected ICLR 2024 reviews.}
    \label{fig:paper_review}
\end{figure*}

\paragraph{Paper review analysis}
To evaluate our multi-agent review system, we conducted a study using the released ICLR 2024 reviews. We randomly selected 50 papers spanning diverse rating levels, and report the results in Figure~\ref{fig:paper_review}. We observe that the code-oriented agent (\texttt{qwen3-coder-30B}) often struggles to sustain a coherent review workflow, whereas chat-style LLMs (e.g., \texttt{gpt-oss}) produce stronger and more consistent reviews. Overall, review quality improves with larger models, suggesting that capacity and instruction-following are particularly important for end-to-end reviewing.

\paragraph{Qualitative assessment}
We evaluated PaperCircle through 81 real-world discovery sessions
(78 unique queries) conducted by researchers across diverse topics. The analysis of the results is shwon in the Table~\ref{tab:retrieval_comparison} and in Table~\ref{tab:papercircle_summary}. The 81 sessions span 9 research domains including world models, LLM training, neural architectures, multi-agent systems, healthcare AI (11\%), model efficiency (10\%), domain-specific applications (10\%), computer vision (7\%), and scientific reasoning (6\%), demonstrating domain-agnostic applicability. The table below compares measurable discovery outcomes against the capabilities of standard single-source search tools.

\begin{table}[ht]
\centering
\caption{Comparison of source coverage and export-related functionality across literature discovery systems. Percentages are computed with respect to the PaperCircle paper set. $\dagger$ Fraction of PaperCircle's 21,115 papers not retrievable from that single source alone.
$\ddagger$ Estimated based on the natural query.}
\label{tab:retrieval_comparison}
\small
\renewcommand{\arraystretch}{1.2}
\setlength{\tabcolsep}{6pt}

\resizebox{\columnwidth}{!}{%
\begin{tabular}{>{\raggedright\arraybackslash}p{3.6cm}cccc}
\toprule
\textbf{Metric} & \textbf{arXiv} & \textbf{Semantic Scholar} & \textbf{Google Scholar} & \textbf{PaperCircle} \\
\midrule
Sources queried per run & 1 & 1 & 1 & \textbf{8.7 avg.} \\
Papers not retrievable$\dagger$ & 70.9\% & 80.4\% & 36.9\%$\ddagger$ & \textbf{9.0\%} \\
PDF availability & \textasciitilde 90\% & \textasciitilde 60\% & Variable & \textbf{62.5\%} \\
Supported export formats & 0 & 1--2 & 1 & \textbf{5} \\
Bulk export support & \xmark & \xmark & \xmark & \textbf{\cmark} \\
Process-level logs & \xmark & \xmark & \xmark & \textbf{\cmark} \\
\bottomrule
\end{tabular}%
}

\end{table}



\begin{table}[ht]
\centering
\caption{Summary statistics of Paper Circle usage and outputs.}
\label{tab:papercircle_summary}
\small
\renewcommand{\arraystretch}{1.2}
\setlength{\tabcolsep}{8pt}
\resizebox{0.5\textwidth}{!}{%
\begin{tabular}{>{\raggedright\arraybackslash}p{3.2cm}
                >{\centering\arraybackslash}p{3cm}
                >{\raggedright\arraybackslash}p{4.8cm}}
\toprule
\textbf{Metric} & \textbf{Value} & \textbf{Interpretation} \\
\midrule
Sessions & 81 & Observed user sessions \\
Papers & 21,115 & Total papers processed \\
arXiv miss & 70.9\% & Fraction not retrievable from arXiv \\
Semantic Scholar miss & 80.4\% & Fraction not retrievable from Semantic Scholar alone \\
Duplicates removed & 18,613 (43.5\%) & Duplicate entries removed during processing \\
Median time & 2.3 min & Median runtime per session \\
Export formats & 5 / session & Number of supported export formats per session \\
\bottomrule
\end{tabular}%
}
\end{table}

Preliminary user feedback indicates minimal cognitive load when using PaperCircle. NASA-TLX~\cite{colligan2015cognitive} assessment yields an overall workload of 1.2/7, with five of six dimensions scoring the minimum (1/7) and effort at 2/7. Usability ratings are correspondingly strong: positive items (frequency of use, ease, integration, learnability, confidence) average 7.6/10, while negative items (complexity, support needs, inconsistency, cumbersomeness, learning curve) average 2.6/10. Notably, the participant rated learnability at 8/10 and learning barrier at 1/10, suggesting the system is accessible without prior training.

\subsection{Ablation Studies}
\label{sec:ablations}

We conduct comprehensive ablation studies to understand the contribution of different system components, including retrieval baselines, query configuration, and pipeline structures.

\paragraph{Full Query utilization}
\label{subsec:full_agent}

To assess the full capability of our system, we conducted an extended evaluation using the \texttt{qwen3-coder-30b} model across 500 queries under various configurations. Results are presented in Table~\ref{tab:qooba_benchmark}.

\begin{table}[htbp]
\centering
\caption{Extended benchmark results for the Qooba agent (\texttt{qwen3-coder-30b}) across different configurations.}
\label{tab:qooba_benchmark}
\resizebox{0.48\textwidth}{!}{
\begin{tabular}{@{}lcccccc@{}}
\toprule
\textbf{Configuration} & \textbf{Queries} & \textbf{Hit Rate} & \textbf{MRR} & \textbf{R@1} & \textbf{R@5} & \textbf{Time (s)} \\
\midrule
Default (Full Agent) & 500 & \textbf{0.9818} & \textbf{0.8824} & \textbf{0.8381} & \textbf{0.9312} & \textbf{21.54} \\
With Filters \& Offline & 50 & 0.9600 & 0.8485 & 0.7800 & 0.9000 & 22.76 \\
Offline Only & 50 & 0.9200 & 0.6476 & 0.5600 & 0.7400 & 41.45 \\
No Mentions & 50 & 0.6400 & 0.4316 & 0.3600 & 0.5200 & 38.35 \\
Online/Offline Mix & 50 & 0.6200 & 0.4595 & 0.4200 & 0.5000 & 38.50 \\
\bottomrule
\end{tabular}
}
\end{table}

\paragraph{Observations.}
 The ``With Filters \& Offline'' configuration performs better, suggesting that explicit context (conference/year filters) combined with local database access is highly effective. Notably, the ``No Mentions'' and ``Online/Offline Mix'' configurations show significant performance degradation (62--64\% hit rate), indicating that specific paper references and structured retrieval chains are critical for accuracy. Overall, configurations exhibit similar latency, indicating stable scaling of the multi-agent pipeline across query settings as well.
\subsection{Retrieval Baseline Ablations}
\label{subsec:retrieval_ablations}


\begin{table}[htbp]
\centering
\caption{Ablation study results comparing retrieval baselines and pipeline structures using \texttt{qwen3-coder-30b}. Full represents the full pipeline structure, minimal represents  }
\label{tab:ablations}
\resizebox{0.48\textwidth}{!}{
\begin{tabular}{@{}llcccccc@{}}
\toprule
\textbf{Configuration} & \textbf{Baseline} & \textbf{Structure} & \textbf{Hit Rate} & \textbf{MRR} & \textbf{R@1} & \textbf{R@5} & \textbf{Time (s)} \\
\midrule
BM25 Full & bm25 & full & 0.9600 & 0.8629 & 0.8000 & 0.9200 & 33.75 \\
BM25 Search Sort & bm25 & search\_sort & 0.9600 & 0.8620 & 0.8000 & 0.9200 & 33.95 \\
BM25 No Intent & bm25 & no\_intent & 0.9600 & 0.8554 & 0.8000 & 0.9200 & \textbf{31.47} \\
BM25 Search Analysis & bm25 & search\_analysis & 0.9600 & 0.8437 & 0.7800 & 0.9200 & 32.81 \\
BM25 Minimal & bm25 & minimal & 0.9600 & 0.8420 & 0.7800 & 0.9200 & 33.34 \\
Hybrid Full & hybrid & full & 0.9600 & 0.8620 & 0.8000 & 0.9200 & 31.65 \\
BM25 + Reranker & bm25+reranker & full & 0.9600 & \textbf{0.8692} & 0.8000 & \textbf{0.9400} & 935.07 \\
Semantic Full & semantic & full & 0.9400 & 0.7097 & 0.6200 & 0.8800 & 31.28 \\
\bottomrule
\end{tabular}
}
\end{table}


\paragraph{Retrieval Baseline Impact.}
BM25-based methods consistently outperform pure semantic retrieval. The semantic baseline shows a significant drop in R@1 (0.62) compared to BM25-based methods (0.80), suggesting that lexical matching remains crucial for precise paper retrieval. The hybrid approach performs on par with BM25, indicating that combining lexical and semantic signals does not provide additional benefits in this setting.

\paragraph{Reranking Trade-offs.}
The BM25 + Reranker configuration achieves the highest MRR (0.8692) and R@5 (0.9400), but at a substantial computational cost, approximately 28$\times$ slower than other methods. This presents a clear accuracy-efficiency trade-off that practitioners must consider based on their deployment requirements.

\paragraph{Pipeline Complexity.}
Reducing pipeline complexity (Minimal, Search Analysis configurations) leads to slight drops in MRR and R@1 while maintaining high overall hit rates (96\%). Interestingly, removing intent analysis (``No Intent'' configuration) results in a faster pipeline with competitive performance, suggesting that intent classification may be redundant for well-structured queries.


\section{Conclusion}
Paper Circle shows how multi-agent workflows can streamline research literature management. Its discovery pipeline unifies heterogeneous search sources and multi-criteria scoring into a reproducible tool, using a simple agent–tool interface with shared state, deterministic ranking, and synchronized multi-format outputs. Its analysis pipeline converts papers into structured knowledge graphs that enable graph-aware QA, coverage checks, and human-in-the-loop verification. Future work will focus on the optimization of the unification of the pipeline.
\section{Limitations}
Our review agent shows weak alignment with human judgments: across models, the correlation with human reviewer scores remains low ($r<0.25$), and several metrics can even exhibit negative correlations, indicating that the system may rank papers in the opposite order of human preference. As a result, even the best-performing configurations do not reliably distinguish strong from weak submissions, and the system should not be used as a trusted mechanism for comparing or ranking papers. Based on our analysis, we found that this review process gets the benefit of a large model, so this problem can be overcome by large open/closed source models. 
\bibliography{main}

@Misc{smolagents,
  title =        {`smolagents`: a smol library to build great agentic systems.},
  author =       {Aymeric Roucher and Albert Villanova del Moral and Thomas Wolf and Leandro von Werra and Erik Kaunismäki},
  howpublished = {\url{https://github.com/huggingface/smolagents}},
  year =         {2025}
}

@article{le2025_0,
  title = {Multi-Agent Causal Discovery Using Large Language Models},
  author = {Hao Duong Le and Xin Xia and Chen Zhang},
  year = {2025},
  journal = {ICLR 2025},
  url = {https://openreview.net/forum?id=Idygh9MX0N},
}

@article{yu2025_1,
  title = {Research Town: Simulator of Research Community},
  author = {Haofei Yu and Zirui Cheng and Zhaochen Hong and Kunlun Zhu and Jinwei Yao and Tao Feng and Jiaxuan You},
  year = {2025},
  journal = {ICLR 2025},
  url = {https://openreview.net/forum?id=IwhvaDrL39},
}

@article{unknown2026_2,
  title={Cellagent: An llm-driven multi-agent framework for automated single-cell data analysis},
  author={Xiao, Yihang and Liu, Jinyi and Zheng, Yan and Xie, Xiaohan and Hao, Jianye and Li, Mingzhi and Wang, Ruitao and Ni, Fei and Li, Yuxiao and Luo, Jintian and others},
  journal={arXiv preprint arXiv:2407.09811},
  year={2024}
}

@article{yu2025_3,
  title = {ResearchTown: Simulator of Human Research Community},
  author = {Haofei Yu and Zhaochen Hong and Zirui Cheng and Kunlun Zhu and Keyang Xuan and Jinwei Yao and Tao Feng and Jiaxuan You},
  year = {2025},
  journal = {ICML 2025},
  url = {https://icml.cc/virtual/2025/poster/46055},
}

@article{unknown2026_5,
  title={Llm-based multi-agent blackboard system for information discovery in data science},
  author={Salemi, Alireza and Parmar, Mihir and Goyal, Palash and Song, Yiwen and Yoon, Jinsung and Zamani, Hamed and Palangi, Hamid and Pfister, Tomas},
  journal={arXiv preprint arXiv:2510.01285},
  year={2025}
}

@article{unknown2026_6,
  title={PiFlow: Principle-aware Scientific Discovery with Multi-Agent Collaboration},
  author={Pu, Yingming and Lin, Tao and Chen, Hongyu},
  journal={arXiv preprint arXiv:2505.15047},
  year={2025}
}

@article{unknown2026_7,
  title={Causal-copilot: An autonomous causal analysis agent},
  author={Wang, Xinyue and Zhou, Kun and Wu, Wenyi and Singh, Har Simrat and Nan, Fang and Jin, Songyao and Philip, Aryan and Patnaik, Saloni and Zhu, Hou and Singh, Shivam and others},
  journal={arXiv preprint arXiv:2504.13263},
  year={2025}
}

@article{chen2025_10,
  title = {AI-Driven Automation Can Become the Foundation of Next-Era Science of Science Research},
  author = {Renqi Chen and Haoyang Su and SHIXIANG TANG and Zhenfei Yin and Qi Wu and Hui Li and Ye Sun and Wanli Ouyang and Philip Torr and Nanqing Dong},
  year = {2025},
  journal = {NIPS 2025},
  url = {https://openreview.net/forum?id=u0FB996GIH},
}

@article{jin2024agentreview,
  title={Agentreview: Exploring peer review dynamics with llm agents},
  author={Jin, Yiqiao and Zhao, Qinlin and Wang, Yiyang and Chen, Hao and Zhu, Kaijie and Xiao, Yijia and Wang, Jindong},
  journal={arXiv preprint arXiv:2406.12708},
  year={2024}
}

@article{d2024marg,
  title={Marg: Multi-agent review generation for scientific papers},
  author={D'Arcy, Mike and Hope, Tom and Birnbaum, Larry and Downey, Doug},
  journal={arXiv preprint arXiv:2401.04259},
  year={2024}
}

@article{das2023comparative,
  title={A comparative study on tf-idf feature weighting method and its analysis using unstructured dataset},
  author={Das, Mamata and Alphonse, PJA and others},
  journal={arXiv preprint arXiv:2308.04037},
  year={2023}
}

@article{zhu2024llms,
  title={Llms for knowledge graph construction and reasoning: Recent capabilities and future opportunities},
  author={Zhu, Yuqi and Wang, Xiaohan and Chen, Jing and Qiao, Shuofei and Ou, Yixin and Yao, Yunzhi and Deng, Shumin and Chen, Huajun and Zhang, Ningyu},
  journal={World Wide Web},
  volume={27},
  number={5},
  pages={58},
  year={2024},
  publisher={Springer}
}

@article{zhang2025schema,
  title={Schema Generation for Large Knowledge Graphs Using Large Language Models},
  author={Zhang, Bohui and He, Yuan and Pintscher, Lydia and Pe{\~n}uela, Albert Mero{\~n}o and Simperl, Elena},
  journal={arXiv preprint arXiv:2506.04512},
  year={2025}
}

@inproceedings{qu2025semantic,
  title={Is semantic chunking worth the computational cost?},
  author={Qu, Renyi and Tu, Ruixuan and Bao, Forrest},
  booktitle={Findings of the Association for Computational Linguistics: NAACL 2025},
  pages={2155--2177},
  year={2025}
}

@article{adhikari2024comparative,
  title={A comparative study of pdf parsing tools across diverse document categories},
  author={Adhikari, Narayan S and Agarwal, Shradha},
  journal={arXiv preprint arXiv:2410.09871},
  year={2024}
}

@article{chen2023bm25,
  title={Bm25 query augmentation learned end-to-end},
  author={Chen, Xiaoyin and Wiseman, Sam},
  journal={arXiv preprint arXiv:2305.14087},
  year={2023}
}

@article{mishra2024granite,
  title={Granite code models: A family of open foundation models for code intelligence},
  author={Mishra, Mayank and Stallone, Matt and Zhang, Gaoyuan and Shen, Yikang and Prasad, Aditya and Soria, Adriana Meza and Merler, Michele and Selvam, Parameswaran and Surendran, Saptha and Singh, Shivdeep and others},
  journal={arXiv preprint arXiv:2405.04324},
  year={2024}
}

@article{hui2024qwen2,
      title={Qwen2. 5-Coder Technical Report},
      author={Hui, Binyuan and Yang, Jian and Cui, Zeyu and Yang, Jiaxi and Liu, Dayiheng and Zhang, Lei and Liu, Tianyu and Zhang, Jiajun and Yu, Bowen and Dang, Kai and others},
      journal={arXiv preprint arXiv:2409.12186},
      year={2024}
}

@article{zhu2024deepseek,
  title={DeepSeek-Coder-V2: Breaking the Barrier of Closed-Source Models in Code Intelligence},
  author={Zhu, Qihao and Guo, Daya and Shao, Zhihong and Yang, Dejian and Wang, Peiyi and Xu, Runxin and Wu, Y and Li, Yukun and Gao, Huazuo and Ma, Shirong and others},
  journal={arXiv preprint arXiv:2406.11931},
  year={2024}
}

@misc{qwen3technicalreport,
      title={Qwen3 Technical Report}, 
      author={Qwen Team},
      year={2025},
      eprint={2505.09388},
      archivePrefix={arXiv},
      primaryClass={cs.CL},
      url={https://arxiv.org/abs/2505.09388}, 
}

@inproceedings{gong2025past,
  title={Past-Future Scheduler for LLM Serving under SLA Guarantees},
  author={Gong, Ruihao and Bai, Shihao and Wu, Siyu and Fan, Yunqian and Wang, Zaijun and Li, Xiuhong and Yang, Hailong and Liu, Xianglong},
  booktitle={Proceedings of the 30th ACM International Conference on Architectural Support for Programming Languages and Operating Systems, Volume 2},
  pages={798--813},
  year={2025}
}

@article{wang2020minilm,
  title={Minilm: Deep self-attention distillation for task-agnostic compression of pre-trained transformers},
  author={Wang, Wenhui and Wei, Furu and Dong, Li and Bao, Hangbo and Yang, Nan and Zhou, Ming},
  journal={Advances in neural information processing systems},
  volume={33},
  pages={5776--5788},
  year={2020}
}

@article{qwen3embedding,
  title={Qwen3 Embedding: Advancing Text Embedding and Reranking Through Foundation Models},
  author={Zhang, Yanzhao and Li, Mingxin and Long, Dingkun and Zhang, Xin and Lin, Huan and Yang, Baosong and Xie, Pengjun and Yang, An and Liu, Dayiheng and Lin, Junyang and Huang, Fei and Zhou, Jingren},
  journal={arXiv preprint arXiv:2506.05176},
  year={2025}
}

@article{han2025deep,
  title={Deep researcher with test-time diffusion},
  author={Han, Rujun and Chen, Yanfei and CuiZhu, Zoey and Miculicich, Lesly and Sun, Guan and Bi, Yuanjun and Wen, Weiming and Wan, Hui and Wen, Chunfeng and Ma{\^\i}tre, Sol{\`e}ne and others},
  journal={arXiv preprint arXiv:2507.16075},
  year={2025}
}

@book{wooldridge2002,
  title={An introduction to multiagent systems},
  author={Wooldridge, Michael},
  year={2002},
  publisher={John Wiley \& Sons}
}

@article{ju2025_16,
  title = {ChemThinker: Thinking Like a Chemist with Multi-Agent LLMs for Deep Molecular Insights},
  author = {Jiaxin Ju and YIZHEN ZHENG and Huan Yee Koh and Can Wang and Shirui Pan},
  year = {2025},
  journal = {ICLR 2025},
  url = {https://openreview.net/forum?id=zlAUnwhE2v},
}

@article{unknown2026_17,
  title={AlphaResearch: Accelerating New Algorithm Discovery with Language Models},
  author={Yu, Zhaojian and Feng, Kaiyue and Zhao, Yilun and He, Shilin and Zhang, Xiao-Ping and Cohan, Arman},
  journal={arXiv preprint arXiv:2511.08522},
  year={2025}
}

@article{unknown2026_19,
  title={Chain-of-agents: End-to-end agent foundation models via multi-agent distillation and agentic rl},
  author={Li, Weizhen and Lin, Jianbo and Jiang, Zhuosong and Cao, Jingyi and Liu, Xinpeng and Zhang, Jiayu and Huang, Zhenqiang and Chen, Qianben and Sun, Weichen and Wang, Qiexiang and others},
  journal={arXiv preprint arXiv:2508.13167},
  year={2025}
}

@article{unknown2026_20,
  title={MARS: Optimizing Dual-System Deep Research via Multi-Agent Reinforcement Learning},
  author={Chen, Guoxin and Qiao, Zile and Wang, Wenqing and Yu, Donglei and Chen, Xuanzhong and Sun, Hao and Liao, Minpeng and Fan, Kai and Jiang, Yong and Zhao, Wayne Xin and others},
  journal={arXiv preprint arXiv:2510.04935},
  year={2025}
}

@article{doru2024_21,
  title = {NegoLog: An Integrated Python-based Automated Negotiation Framework with Enhanced Assessment Components},
  author = {Anıl Doğru and Mehmet Onur Keskin and Catholijn M. Jonker and Tim Baarslag and Reyhan Aydoğan},
  year = {2024},
  journal = {IJCAI 2024},
  url = {https://www.ijcai.org/proceedings/2024/998},
}

@article{lala2023paperqa,
    title = {PaperQA: Retrieval-Augmented Generative Agent for Scientific Research},
    author = {
    Jakub Lála and
 Odhran O'Donoghue and
 Aleksandar Shtedritski and
 Sam Cox and
 Samuel G. Rodriques and
 Andrew D. White},
    journal = {arXiv preprint arXiv:2312.07559},
    year = {2023},
    url = {https://doi.org/10.48550/arXiv.2312.07559}
}

@article{colligan2015cognitive,
  title={Cognitive workload changes for nurses transitioning from a legacy system with paper documentation to a commercial electronic health record},
  author={Colligan, Lacey and Potts, Henry WW and Finn, Chelsea T and Sinkin, Robert A},
  journal={International journal of medical informatics},
  volume={84},
  number={7},
  pages={469--476},
  year={2015},
  publisher={Elsevier}
}

@inproceedings{shao-etal-2024-assisting,
    title = "Assisting in Writing {W}ikipedia-like Articles From Scratch with Large Language Models",
    author = "Shao, Yijia  and
      Jiang, Yucheng  and
      Kanell, Theodore  and
      Xu, Peter  and
      Khattab, Omar  and
      Lam, Monica",
    editor = "Duh, Kevin  and
      Gomez, Helena  and
      Bethard, Steven",
    booktitle = "Proceedings of the 2024 Conference of the North American Chapter of the Association for Computational Linguistics: Human Language Technologies (Volume 1: Long Papers)",
    month = jun,
    year = "2024",
    address = "Mexico City, Mexico",
    publisher = "Association for Computational Linguistics",
    url = "https://aclanthology.org/2024.naacl-long.347/",
    doi = "10.18653/v1/2024.naacl-long.347",
    pages = "6252--6278",
}

@article{shi2025scisage,
  title={Scisage: A multi-agent framework for high-quality scientific survey generation},
  author={Shi, Xiaofeng and Kou, Qian and Li, Yuduo and Tang, Ning and Xie, Jinxin and Yu, Longbin and Wang, Songjing and Zhou, Hua},
  journal={arXiv preprint arXiv:2506.12689},
  year={2025}
}

@article{
unknown2026_22,
  title={Reinforce LLM Reasoning through Multi-Agent Reflection},
  author={Yuan, Yurun and Xie, Tengyang},
  journal={arXiv preprint arXiv:2506.08379},
  year={2025}
}

@article{yang2025_25,
  title = {MOOSE-Chem: Large Language Models for Rediscovering Unseen Chemistry Scientific Hypotheses},
  author = {Zonglin Yang and Wanhao Liu and Ben Gao and Tong Xie and Yuqiang Li and Wanli Ouyang and Soujanya Poria and Erik Cambria and Dongzhan Zhou},
  year = {2025},
  journal = {ICLR 2025},
  url = {https://iclr.cc/virtual/2025/poster/29319},
}

@article{unknown2026_26,
  title={ChemBOMAS: Accelerated BO in Chemistry with LLM-Enhanced Multi-Agent System},
  author={Han, Dong and Ai, Zhehong and Cai, Pengxiang and Lu, Shanya and Chen, Jianpeng and Ye, Zihao and Sun, Shuzhou and Gao, Ben and Ge, Lingli and Wang, Weida and others},
  journal={arXiv preprint arXiv:2509.08736},
  year={2025}
}

@article{chu2024_27,
  title = {Cohesive Conversations: Enhancing Authenticity in Multi-Agent Simulated Dialogues},
  author = {KuanChao Chu and Yi-Pei Chen and Hideki Nakayama},
  year = {2024},
  journal = {COLM 2024},
  url = {https://openreview.net/forum?id=3ypWPhMGhV},
}

@article{reddy2025_30,
  title = {Towards Scientific Discovery with Generative AI: Progress, Opportunities, and Challenges},
  author = {Chandan K Reddy and Parshin Shojaee},
  year = {2025},
  journal = {AAAI 2025},
  url = {https://ojs.aaai.org/index.php/AAAI/article/view/35084},
}

@article{keskin2024_31,
  title = {NEGOTIATOR: A Comprehensive Framework for Human-Agent Negotiation Integrating Preferences, Interaction, and Emotion},
  author = {Mehmet Onur Keskin and Berk Buzcu and Berkecan Koçyiğit and Umut Çakan and Anıl Doğru and Reyhan Aydoğan},
  year = {2024},
  journal = {IJCAI 2024},
  url = {https://www.ijcai.org/proceedings/2024/1012},
}

@article{xiao2025_34,
  title = {MoodAngels: A Retrieval-augmented Multi-agent Framework for Psychiatry Diagnosis},
  author = {Mengxi Xiao and Ben Liu and He Li and Jimin Huang and Qianqian Xie and Xiaofen Zong and Mang Ye and Min Peng},
  year = {2025},
  journal = {NIPS 2025},
  url = {https://openreview.net/forum?id=AWU93F6Bup},
}

@article{kumbhar2025_37,
  title = {Hypothesis Generation for Materials Discovery and Design Using Goal-Driven and Constraint-Guided LLM Agents},
  author = {Shrinidhi Kumbhar and Venkatesh Mishra and Kevin Coutinho and Divij Handa and Ashif Iquebal and Chitta Baral},
  year = {2025},
  journal = {NAACL 2025},
  url = {https://aclanthology.org/2025.findings-naacl.420/},
}

@article{pramanick2023_39,
  title = {A Diachronic Analysis of Paradigm Shifts in NLP Research: When, How, and Why?},
  author = {Aniket Pramanick and Yufang Hou and Saif M. Mohammad and Iryna Gurevych},
  year = {2023},
  journal = {EMNLP 2023},
  url = {https://openreview.net/forum?id=qhwYFIrSm7},
}

@article{chandrashekar2025_47,
  title = {A SURVEY ON STOCK INVESTMENT RISK ANALYSIS USING CREWAI MULTI- AGENT SYSTEM},
  author = {Prof. Chandrashekar and M. Akram and Mohin Khan and Piyush Kumar and Pratap Mandal},
  year = {2025},
  journal = {International Research Journal of Modernization in Engineering Technology and Science},
  doi = {10.56726/irjmets66945},
  url = {https://www.semanticscholar.org/paper/f17c5b4bc435f7ad7714290797036250717bcaec},
}

@article{wawer2025_50,
  title = {Integrating Traditional Technical Analysis with AI: A Multi-Agent LLM-Based Approach to Stock Market Forecasting},
  author = {Michał Wawer and Jarosław A. Chudziak},
  year = {2025},
  journal = {International Conference on Agents and Artificial Intelligence},
  doi = {10.5220/0013191200003890},
  url = {https://www.semanticscholar.org/paper/d96ce80541b552fd703291594939bc9d624bb7ae},
}

@article{naumov2025_51,
  title = {DORA AI Scientist: Multi-agent Virtual Research Team for Scientific Exploration Discovery and Automated Report Generation},
  author = {Vladimir Naumov and Diana Zagirova and Sha Lin and Yupeng Xie and Wenhao Gou and Anatoly Urban and Nina Tikhonova and Khadija M. Alawi and Mike Durymanov and Fedor Galkin},
  year = {2025},
  journal = {bioRxiv},
  doi = {10.1101/2025.03.06.641840},
  url = {https://www.semanticscholar.org/paper/b913baa47245a4a76054bb12e04d61c3c88fb532},
}

@article{niyakan2025_52,
  title = {PhenoGraph: A Multi-Agent Framework for Phenotype-driven Discovery in Spatial Transcriptomics Data Augmented with Knowledge Graphs},
  author = {Seyednami Niyakan and Xiaoning Qian},
  year = {2025},
  journal = {bioRxiv},
  doi = {10.1101/2025.06.06.658341},
  url = {https://www.semanticscholar.org/paper/9dc71ca68ad2977dbebbcf6137f198a41ffb4ee2},
}

@article{spieser2025_53,
  title = {Multi-Agent AI Systems for Biological and Clinical Data Analysis},
  author = {Jackson Spieser and Ali Balapour and Jarek Meller and Krushna Patra and Behrouz Shamsaei},
  year = {2025},
  journal = {Preprints.org},
  doi = {10.20944/preprints202512.2602.v1},
  url = {https://openalex.org/W7117730344},
}

@article{fehlis2025_56,
  title = {Accelerating Drug Discovery Through Agentic AI: A Multi-Agent Approach to Laboratory Automation in the DMTA Cycle},
  author = {Yao Fehlis and Charles Crain and Aidan Jensen and Michael Watson and James Juhasz and Paul Mandel and Betty Liu and Shawn Mahon and Daren Wilson and Nick Lynch-Jonely},
  year = {2025},
  journal = {arXiv.org},
  doi = {10.48550/arXiv.2507.09023},
  url = {https://www.semanticscholar.org/paper/7778c3dc1ca422cd87c6482cfc451a29ec941e5f},
}

@article{singh2025_57,
  title = {ASTRAFIN:- AI Financial Agent},
  author = {Er. Jagpreet Singh and Prasant Kumar},
  year = {2025},
  journal = {INTERNATIONAL JOURNAL OF SCIENTIFIC RESEARCH IN ENGINEERING AND MANAGEMENT},
  doi = {10.55041/ijsrem54152},
  url = {https://www.semanticscholar.org/paper/27124a1778b69afd77b169c4e5145fbee8706040},
}

@article{korat2025_58,
  title = {Synergistic minds: A collaborative multi-agent framework for integrated AI tool development using diverse large language models},
  author = {Arpan Shaileshbhai Korat},
  year = {2025},
  journal = {World Journal of Advanced Research and Reviews},
  doi = {10.30574/wjarr.2025.27.2.1806},
  url = {https://www.semanticscholar.org/paper/be235a38449b7d489af5f6a583b734979fe2100d},
}

@article{gajjar2025_61,
  title = {EvoResearch: A Multi-Agent AI Framework for Automated Paper Analysis},
  author = {Prof.Anjali Gajjar},
  year = {2025},
  journal = {International Journal of Innovative Research in Advanced Engineering},
  doi = {10.26562/ijirae.2025.v1211.25},
  url = {https://www.semanticscholar.org/paper/d05a5180798bbbdc411aad66e441c053b382ea44},
}

\appendix

\section{Paper Review Results}
We evaluate how well large language models can predict human paper-review scores on ICLR submissions.
From the ICLR 2024 dataset, we randomly sampled 50 papers to cover a broad range of human-assigned ratings and evaluated four tool-enabled LLMs: \texttt{gpt-oss:120b}, \texttt{gpt-oss:20b}, \texttt{qwen3-coder-30b}, and a quantized \texttt{qwen3-coder-30b} variant.
For each paper, the model produces numerical scores for standard review dimensions (overall rating, soundness, presentation, and contribution), which we compare against the corresponding human scores.

\paragraph{Metrics.}
We report regression error (MSE, MAE, RMSE), rank/linear association (Pearson, Spearman), and thresholded accuracy (percentage of predictions within $\pm 0.5$, $\pm 1.0$, and $\pm 1.5$ of the human score).
We also report the mean and standard deviation of signed errors to characterize systematic bias.
Due to occasional missing fields or filtering during preprocessing, the number of evaluated papers $N$ can differ slightly across models.

\paragraph{Key findings.}
Across categories, \texttt{gpt-oss:120b} achieves the best overall accuracy on \textit{rating} and \textit{contribution} (e.g., rating MAE $=1.68$; contribution MAE $=0.62$), while \texttt{gpt-oss:20b} is competitive and often stronger on more technical sub-scores such as \textit{soundness} and \textit{presentation}.
Despite moderate absolute errors on several dimensions, correlations with human scores remain weak across models (generally $|r|<0.25$), suggesting that models struggle to preserve the relative ranking of papers even when their average deviation is limited.
Code-specialized models (Qwen3-Coder) remain viable baselines, but show larger errors on overall rating and contribution in this setting.

\begin{table*}[t]
\centering
\renewcommand{\arraystretch}{1.15}
\resizebox{0.9\textwidth}{!}{
\begin{tabular}{llrrrrrrrrrrr}
\hline
\textbf{Model} & \textbf{Category} &
\textbf{MSE} & \textbf{MAE} & \textbf{RMSE} &
\textbf{Pearson} & \textbf{Spearman} &
\textbf{Acc. $\pm0.5$} & \textbf{Acc. $\pm1.0$} & \textbf{Acc. $\pm1.5$} &
\textbf{Mean Err.} & \textbf{Std Err.} & \textbf{N} \\
\hline
oss-120B & RATING        & 4.6934 & 1.6844 & 2.1664 & -0.0407 &  0.0571 & 25.00\% & 43.75\% & 58.33\% &  0.2177 & 2.1555 & 48 \\
oss-120B & SOUNDNESS     & 0.7316 & 0.6351 & 0.8554 & -0.0054 &  0.0474 & 58.33\% & 85.42\% & 87.50\% & -0.0816 & 0.8515 & 48 \\
oss-120B & PRESENTATION  & 0.6564 & 0.6038 & 0.8102 &  0.0701 &  0.1259 & 60.42\% & 83.33\% & 91.67\% & -0.0920 & 0.8049 & 48 \\
oss-120B & CONTRIBUTION  & 0.6349 & 0.6240 & 0.7968 &  0.0717 &  0.0734 & 56.25\% & 85.42\% & 91.67\% &  0.0087 & 0.7967 & 48 \\
\hline
oss-20 & RATING          & 4.7607 & 1.7647 & 2.1819 &  0.0989 &  0.1869 & 21.43\% & 40.48\% & 52.38\% &  1.5980 & 1.4856 & 42 \\
oss-20 & SOUNDNESS       & 0.4241 & 0.5190 & 0.6512 & -0.0106 & -0.0226 & 59.52\% & 92.86\% & 97.62\% &  0.3294 & 0.5618 & 42 \\
oss-20 & PRESENTATION    & 0.4271 & 0.5171 & 0.6535 & -0.1270 & -0.1299 & 64.29\% & 90.48\% & 97.62\% &  0.3512 & 0.5511 & 42 \\
oss-20 & CONTRIBUTION    & 0.6482 & 0.6702 & 0.8051 &  0.2221 &  0.1757 & 50.00\% & 83.33\% & 97.62\% &  0.6250 & 0.5075 & 42 \\
\hline
qwen30B-code\_qk\_3 & RATING       & 11.8533 & 2.9879 & 3.4429 & -0.2233 & -0.2837 &  8.51\% & 17.02\% & 29.79\% & 2.9085 & 1.8422 & 47 \\
qwen30B-code\_qk\_3 & SOUNDNESS    &  1.6941 & 1.1730 & 1.3016 &  0.0113 & -0.0096 & 17.02\% & 46.81\% & 72.34\% & 1.1454 & 0.6182 & 47 \\
qwen30B-code\_qk\_3 & PRESENTATION &  1.4257 & 1.0191 & 1.1940 &  0.0378 &  0.0271 & 27.66\% & 59.57\% & 78.72\% & 0.9787 & 0.6840 & 47 \\
qwen30B-code\_qk\_3 & CONTRIBUTION &  2.2921 & 1.3865 & 1.5140 &  0.0196 &  0.0224 & 12.77\% & 34.04\% & 65.96\% & 1.3865 & 0.6080 & 47 \\
\hline
Qwen 30B & RATING         & 10.2331 & 2.7930 & 3.1989 & -0.1820 & -0.2216 &  7.89\% & 13.16\% & 26.32\% & 2.6930 & 1.7266 & 38 \\
Qwen 30B & SOUNDNESS      &  1.7172 & 1.2096 & 1.3104 & -0.1157 & -0.1057 & 13.16\% & 39.47\% & 73.68\% & 1.1491 & 0.6298 & 38 \\
Qwen 30B & PRESENTATION   &  0.9526 & 0.7180 & 0.9760 & -0.1319 & -0.1495 & 55.26\% & 73.68\% & 81.58\% & 0.6522 & 0.7261 & 38 \\
Qwen 30B & CONTRIBUTION   &  2.5212 & 1.4746 & 1.5878 & -0.2119 & -0.2160 & 13.16\% & 26.32\% & 55.26\% & 1.4640 & 0.6146 & 38 \\
\hline
\end{tabular}
}
\caption{Paper review score prediction on ICLR 2024. We compare four LLMs on predicting human review scores across rating, soundness, presentation, and contribution. We report error metrics (MSE/MAE/RMSE), correlation (Pearson/Spearman), and thresholded accuracy (within $\pm 0.5$, $\pm 1.0$, $\pm 1.5$ of the human score). $N$ denotes the number of papers evaluated for each model after preprocessing.}
\label{tab:eval-metrics}
\end{table*}
\section{System Overview}
\textls[-12]{Paper Circle is a full-stack platform with a web frontend and a Python backend as shown in the Figure \ref{fig:front_clint}. The frontend (React, TypeScript, Vite, TailwindCSS) provides discovery, reading circles, and discussion features. The backend exposes discovery APIs via FastAPI and implements the multi-agent pipelines used by the system. Supabase (PostgreSQL + Auth) provides storage for users, communities, papers, and sessions.}

The discovery backend includes two major pipelines: (i) a refactored research discovery pipeline focused on deterministic retrieval, scoring, and diversity, and (ii) a multi-agent research pipeline that produces structured step-by-step outputs with offline search support. Both pipelines are accessible through API endpoints and are integrated into the Paper Circle user interface for interactive discovery workflows.

Figure~\ref{fig:discovery_front} illustrates the overall architecture of Paper Circle. The system consists of two complementary multi-agent pipelines: the \textit{Discovery Pipeline} for finding relevant papers, and the \textit{Analysis Pipeline} for deep understanding of individual papers.

The discovery pipeline, as shown in the Figure \ref{fig:discovery_front} is composed of six agents: intent classification, paper search, sorting, analysis, export, and web search. The intent classifier parses natural-language queries into structured constraints (search mode, conferences, year range, max results, and ranking preferences). The paper search agent is the primary retrieval worker; it updates the global state and writes outputs after every search step. The sorting and analysis agents operate on the shared paper list to refine ranking and derive insights. The export agent centralizes output access for downstream workflows, while the web search agent supplements the pipeline with external lookup tools when required. All agents are coordinated by the \texttt{CodeAgent}, which enforces a minimal-step policy for efficiency and uses the intent classifier to decide offline versus online search.

The analysis pipeline operates on individual papers, transforming PDF documents into structured knowledge graphs. It employs four specialized extraction agents (concept, method, experiment, and linkage) that process paper content in phases, building a typed graph with full traceability to source locations. The resulting graph supports question answering, coverage verification, and multi-format export.

\subsection{State Management and Outputs}
State is maintained in \texttt{PipelineState}. Each step increments a counter, logs action metadata, and regenerates synchronized artifacts. The outputs include: (i) \texttt{papers.json} with full paper metadata and computed scores, (ii) \texttt{links.json} with structured links and PDFs/DOIs, (iii) \texttt{stats.json} with aggregate statistics and a leaderboard, (iv) \texttt{summary.json} with generated insights and key findings, (v) \texttt{retrieval\_metrics.json} when evaluation is enabled, and (vi) human-readable exports (CSV, BibTeX, Markdown) plus a live HTML dashboard. This approach ensures that each agent step is reproducible and auditable.

\subsection{Retrieval}
The pipeline supports both offline and online retrieval. Offline search loads papers from a local JSON corpus and optionally filters by conference and year. It ranks results using BM25 by default, with optional semantic similarity (sentence transformers) or hybrid scoring when available. An optional cross-encoder reranker can refine the top results; when enabled, it reranks a first-stage candidate set. Online search aggregates results from arXiv, Semantic Scholar, OpenAlex, and DBLP via their public APIs. A query intent classifier detects search mode, conference constraints, year ranges, and ranking preferences, and routes the query to the appropriate retrieval pathway. Deduplication is applied across sources by normalizing titles.

\subsection{Ranking and Scoring}
After retrieval, papers are scored along multiple axes: recency, similarity to the query (TF--IDF~\cite{das2023comparative} when available), novelty based on title token frequency, and normalized BM25 scores~\cite{chen2023bm25}. The system supports sorting by any single criterion or by a weighted combined score. Relevance scores are computed as a weighted mixture of similarity, recency, citation count, and BM25. Final ranks are assigned after sorting, and the updated ordering is reflected in all exported artifacts. When reranker-based sorting is requested, a cross-encoder replaces the default scoring with direct relevance scores.

\subsection{Analysis and Monitoring}
The pipeline computes aggregate statistics such as source distribution, year distribution, top authors and venues, keyword frequency, and citation summaries. These analytics populate structured summaries and are visualized in an auto-refreshing HTML dashboard. Each agent action is logged with timestamps and paper counts, enabling reproducibility and step-level auditing of the pipeline. The pipeline also maintains a step log that captures the agent name, action, results preview, and parameters used.

\begin{figure*}
    \centering
    \includegraphics[width=\linewidth]{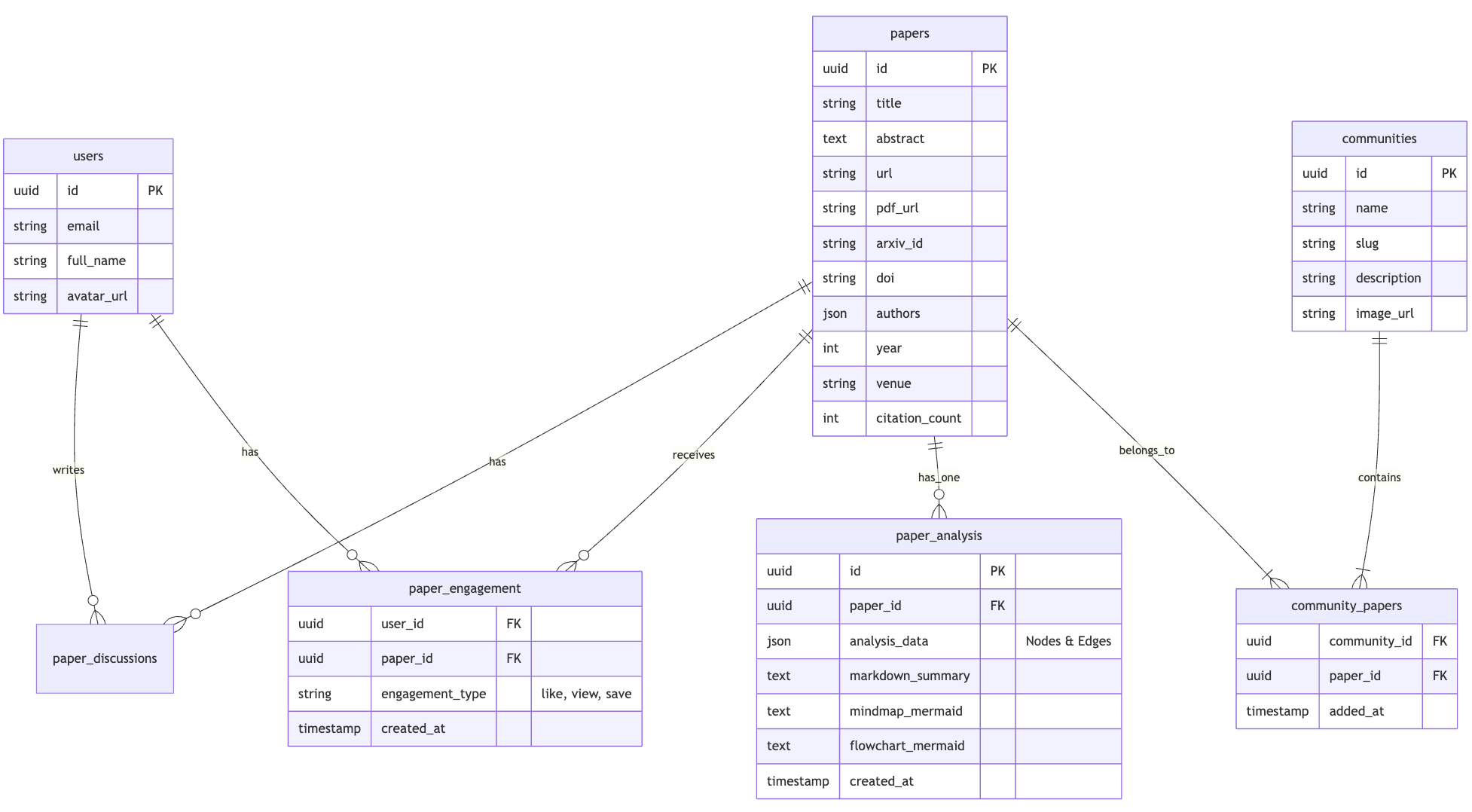}
    \caption{Paper analysis and database management for fast inference.}
    \label{fig:paper_analysis_dbm}
\end{figure*}
\section{Retrieval Pipeline}

Paper Circle supports both offline and online retrieval to balance coverage, speed, and reproducibility. The choice between retrieval modes is controlled by the intent classification agent, which parses user queries to determine the optimal search strategy.

\subsection{Offline Retrieval}

The \texttt{OfflinePaperSearchEngine} enables fast (See the Figure \ref{fig:paper_analysis_dbm}, reproducible search over a local database of academic papers stored as JSON files. Each database file contains structured paper metadata including title, authors, abstract, venue, year, track, keywords, and DOI.

The offline search process:
\begin{enumerate}
    \item \textbf{Database Loading}: Papers are loaded from the specified database path with optional filtering by conference (e.g., ICLR, NeurIPS, ACL) and year range.
    \item \textbf{Text Preparation}: For each paper, searchable text is constructed by concatenating the title, abstract, and keywords.
    \item \textbf{BM25 Indexing}: When available, papers are indexed using the Okapi BM25 algorithm via the \texttt{rank\_bm25} library. The index uses tokenized documents for sparse retrieval.
    \item \textbf{Query Execution}: User queries are tokenized and scored against the BM25 index, returning a ranked list of candidates.
\end{enumerate}

An optional cross-encoder reranker can refine the top-$k$ results from the first-stage retrieval. When enabled via the \texttt{AdvancedReranker} module, the system uses a transformer-based reranker (e.g., Qwen3-Reranker) to compute more precise relevance scores between the query and candidate documents.

\subsection{Online Retrieval}

For broader or more current searches, Paper Circle aggregates results from multiple academic APIs:

\begin{itemize}
    \item \textbf{arXiv}: Queries the arXiv API for preprints, extracting title, authors, abstract, categories, and PDF links.
    \item \textbf{Semantic Scholar}: Retrieves papers with citation counts, abstracts, and venue information via the Semantic Scholar Academic Graph API.
    \item \textbf{OpenAlex}: Accesses the OpenAlex catalog for open-access metadata and citation networks.
    \item \textbf{DBLP}: Searches the DBLP computer science bibliography for venue-specific results.
\end{itemize}

Each source is queried in parallel using a thread pool executor for efficiency. Results are normalized into the common \texttt{Paper} data structure before merging.

\subsection{Deduplication}

After retrieval, the pipeline performs two-stage deduplication to eliminate redundant entries:

\begin{enumerate}
    \item \textbf{DOI-based deduplication}: Papers with matching DOIs are deduplicated, preferring entries with richer metadata (e.g., abstracts, PDF URLs).
    \item \textbf{Title-based deduplication}: Titles are normalized by removing punctuation and converting to lowercase. Duplicate titles are merged, again preferring metadata-complete entries.
\end{enumerate}

The deduplication step is critical when aggregating results from multiple sources, as the same paper often appears in arXiv, Semantic Scholar, and OpenAlex with varying metadata quality.

\subsection{Query Expansion}

The query generation agent converts natural-language user input into a structured search specification containing:
\begin{itemize}
    \item \textbf{Core keywords}: Primary search terms extracted from the query.
    \item \textbf{Required constraints}: Mandatory terms that must appear in results.
    \item \textbf{Related terms}: Synonyms or related concepts to expand recall.
    \item \textbf{Negative keywords}: Terms to exclude from results.
    \item \textbf{Plausible paper titles}: Hypothesized titles for targeted retrieval.
\end{itemize}

This structured specification enables consistent query construction across heterogeneous data sources while capturing user intent more precisely than raw keyword matching.

\section{Scoring and Ranking}

Paper Circle employs a multi-criteria scoring framework designed for research discovery rather than general information retrieval. Each paper receives scores along multiple dimensions, which are combined using mode-specific weights to produce a final ranking.

\subsection{Scoring Dimensions}

The system computes the following scores for each retrieved paper:

\paragraph{Similarity Score} Relevance to the user query is computed using TF--IDF~\cite{das2023comparative} vectorization and cosine similarity. The query and paper text (concatenated title and abstract) are transformed into TF--IDF vectors using scikit-learn's \texttt{TfidfVectorizer}. The similarity score is the cosine of the angle between these vectors:
\begin{equation}
    \text{similarity}(q, p) = \frac{\vec{v}_q \cdot \vec{v}_p}{\|\vec{v}_q\| \cdot \|\vec{v}_p\|}
\end{equation}
where $\vec{v}_q$ and $\vec{v}_p$ are the TF--IDF vectors for the query and paper, respectively.

\paragraph{Recency Score} Papers are scored by publication year, with more recent papers receiving higher scores. The recency score is normalized relative to the current year:
\begin{equation}
    \text{recency}(p) = \frac{\text{year}(p) - \text{year}_{\min}}{\text{year}_{\max} - \text{year}_{\min}}
\end{equation}
where $\text{year}_{\min}$ and $\text{year}_{\max}$ are the minimum and maximum years in the corpus.

\paragraph{Novelty Score} Novelty measures how different a paper is from the corpus centroid, computed as the TF--IDF distance from the average document vector. Papers with unusual terminology or unique topic combinations receive higher novelty scores, surfacing potentially overlooked works.

\paragraph{BM25 Score} When the \texttt{rank\_bm25} library is available, the Okapi BM25 algorithm provides an alternative relevance measure that accounts for term frequency saturation and document length normalization. BM25 scores are normalized to the $[0, 1]$ range for comparability with other dimensions.

\paragraph{Citation Count} When available from the source API (primarily Semantic Scholar and OpenAlex), citation counts provide a proxy for impact. Citation-based ranking is optional and disabled by default to avoid recency bias against new papers.

\subsection{Combined Score Computation}

The final combined score is a weighted sum of individual dimensions:
\begin{equation}
    \text{combined}(p) = w_s \cdot \text{similarity} + w_r \cdot \text{recency} + w_n \cdot \text{novelty} + w_b \cdot \text{bm25}
\end{equation}

The weights $(w_s, w_r, w_n, w_b)$ are determined by the search mode:

\begin{itemize}
    \item \textbf{Stable mode}: Prioritizes relevance and authority. Weights: $w_s = 0.5$, $w_r = 0.2$, $w_n = 0.1$, $w_b = 0.2$.
    \item \textbf{Discovery mode}: Prioritizes novelty to surface non-obvious results. Weights: $w_s = 0.3$, $w_r = 0.1$, $w_n = 0.4$, $w_b = 0.2$.
    \item \textbf{Balanced mode}: Equal emphasis across dimensions. Weights: $w_s = 0.3$, $w_r = 0.2$, $w_n = 0.2$, $w_b = 0.3$.
\end{itemize}

Users can override these weights at query time via API parameters, enabling custom relevance trade-offs for specific research contexts.

\subsection{Sorting Stage}

After scoring, the sorting agent reorders papers according to user preferences. Supported sort criteria include:
\begin{itemize}
    \item \texttt{recency}: Most recent papers first.
    \item \texttt{citations}: Highest-cited papers first.
    \item \texttt{similarity}: Most relevant papers first.
    \item \texttt{novelty}: Most unusual papers first.
    \item \texttt{bm25}: Best BM25 matches first.
    \item \texttt{combined}: Weighted combined score (default).
\end{itemize}

\subsection{Cross-Encoder Reranking}

For high-precision use cases, the pipeline supports optional cross-encoder reranking. When enabled, a transformer-based reranker (configured via \texttt{RerankerConfig}) processes query-document pairs through a cross-attention model to compute more accurate relevance scores than first-stage retrieval alone. The \texttt{MultiStageRetriever} first retrieves a larger candidate set (e.g., top-200) using BM25, then reranks to produce the final top-$k$ results. This two-stage approach balances efficiency with ranking quality.

\section{Diversity and Postprocessing}

Relevance-based ranking alone can produce homogeneous results, with multiple papers covering similar topics or methods. Paper Circle addresses this through diversity-aware postprocessing that ensures the top results span a broader range of perspectives.

\subsection{Maximal Marginal Relevance}

To improve topical coverage, Paper Circle applies Maximal Marginal Relevance (MMR) to the candidate list after initial scoring. MMR iteratively selects papers that maximize a combination of relevance to the query and dissimilarity to already-selected papers:

\begin{equation}
    \text{MMR} = \arg\max_{p \in R \setminus S} \left[ \lambda \cdot \text{sim}(p, q) - (1-\lambda) \cdot \max_{s \in S} \text{sim}(p, s) \right]
\end{equation}

where $R$ is the candidate set, $S$ is the set of already-selected papers, $q$ is the query, and $\lambda$ controls the relevance--diversity trade-off.

The diversity parameter $\lambda$ is mode-dependent:
\begin{itemize}
    \item \textbf{Stable mode}: $\lambda = 0.8$ (relevance-focused).
    \item \textbf{Discovery mode}: $\lambda = 0.5$ (diversity-focused).
    \item \textbf{Balanced mode}: $\lambda = 0.65$.
\end{itemize}

Similarity between papers is computed using TF--IDF cosine similarity over concatenated title and abstract text. This ensures that top results cover distinct subtopics rather than repeating variations of the same idea.

\subsection{Secondary Views}

The pipeline constructs specialized views over the ranked list to serve different discovery goals:

\paragraph{Hidden Gems} Papers with high novelty scores but moderate relevance scores are surfaced as ``hidden gems.'' These are papers that may not rank highly on traditional relevance metrics but offer unique perspectives or cover underexplored topics. The hidden gems view is computed by sorting papers by novelty score and filtering for those below rank 20 in the combined ranking.

\paragraph{Canonical Papers} Papers with high citation counts or appearing in top-tier venues are flagged as ``canonical'' works. This view helps users identify foundational papers in a research area, complementing the recency-focused main ranking.

\paragraph{Source Distribution} The postprocessing stage also reports the distribution of papers across sources (arXiv, Semantic Scholar, etc.), enabling users to assess coverage and identify potential gaps in the retrieval.

\subsection{Statistics and Analytics}

After ranking, the analysis agent computes aggregate statistics stored in \texttt{stats.json}:

\begin{itemize}
    \item \textbf{Year distribution}: Paper counts by publication year.
    \item \textbf{Source distribution}: Paper counts by retrieval source.
    \item \textbf{Top authors}: Authors appearing most frequently in results.
    \item \textbf{Top venues}: Conferences and journals with highest representation.
    \item \textbf{Keyword frequency}: Most common terms in paper titles.
    \item \textbf{Citation statistics}: Total, average, median, min, and max citation counts.
    \item \textbf{Score statistics}: Average similarity, novelty, recency, and BM25 scores.
\end{itemize}

These analytics are visualized in an auto-refreshing HTML dashboard that updates every 10 seconds during pipeline execution, providing real-time visibility into the discovery process.

\subsection{Insight Generation}

The pipeline automatically generates human-readable insights from the collected data:

\begin{itemize}
    \item \textbf{Publication trends}: Identifies the year with the most publications.
    \item \textbf{Primary source}: Reports which API contributed the most results.
    \item \textbf{Prolific authors}: Highlights researchers with multiple papers in the collection.
    \item \textbf{Citation leaders}: Identifies the most-cited paper.
    \item \textbf{Hot topics}: Lists the most frequent keywords.
    \item \textbf{Open access availability}: Reports the percentage of papers with direct PDF links.
\end{itemize}

These insights are stored in \texttt{summary.json} and displayed on the dashboard, helping users quickly understand the landscape of retrieved literature.

\section{Outputs and Interfaces}
The pipeline maintains synchronized structured outputs after every agent step. The primary artifacts include:
\begin{itemize}
  \item \texttt{papers.json}: Full paper metadata and scores.
  \item \texttt{links.json}: Structured links and PDF/DOI entries.
  \item \texttt{stats.json}: Aggregate statistics and leaderboards.
  \item \texttt{summary.json}: Insights and key findings.
  \item \texttt{retrieval\_metrics.json}: Step-level evaluation metrics.
\end{itemize}
Additional exports include CSV, BibTeX, Markdown, and an auto-refreshing HTML dashboard. These outputs allow the same discovery session to be used for curation, citation management, and reporting.

The system exposes REST APIs via FastAPI. The discovery endpoint accepts a query and mode, returns structured search specifications, and provides the full ranked list with scores. Mode weights can be queried or overridden at runtime, enabling customized relevance/authority/novelty trade-offs.

\section{Evaluation}
We evaluate Paper Circle along three axes: (i) retrieval effectiveness under different configurations, (ii) stability and reproducibility of rankings across steps, and (iii) the utility of diversity-aware postprocessing for surfacing non-redundant results. Paper Circle provides built-in evaluation metrics but does not enforce a fixed benchmark dataset. When a ground-truth paper title or identifier is provided, the system computes Mean Reciprocal Rank (MRR), Recall@K, Precision@K, and hit rates. These metrics are computed per step and stored in JSON file for longitudinal tracking.

As a minimal illustrative scenario, consider a known target paper in the local corpus: the pipeline is run once using offline retrieval and once using online sources. The resulting MRR and Recall@K values allow direct comparison of configuration impact, while repeated runs confirm stable rankings when deterministic scoring is enabled. Although lightweight, this framing aligns evaluation with discovery goals rather than task-specific QA benchmarks.

For batch evaluation, a parallel benchmarking utility executes multiple queries concurrently and aggregates mean metrics and timing statistics. This supports lightweight comparisons between search configurations (offline vs. online, BM25 vs. semantic, with or without reranking) without requiring external tooling.

\paragraph{Knowledge Graph Schema.}
The mind graph follows a typed schema with nodes for papers, sections, concepts, methods, experiments, datasets, and visual elements (figures, tables, equations), and edges encoding structural and semantic relations such as hierarchy, definition, proposal, usage, evaluation, illustration, and dependency. Each node and edge is annotated with provenance metadata, including source chunk IDs, page numbers, verification status, confidence scores, and timestamps, providing full traceability from any graph element back to the original PDF.
\subsection{Multi-Agent Extraction}

The \texttt{GraphBuilder} orchestrates four specialized extraction agents, each implemented as a \texttt{CodeAgent} with domain-specific instructions:

\paragraph{Concept Extractor} Identifies key concepts from text chunks, classifying each by type (definition, technique, theory, phenomenon) and importance (core, supporting, background). The agent outputs structured JSON with concept names, descriptions, and classifications.

\paragraph{Method Extractor} Focuses on sections containing method-related keywords (``method'', ``approach'', ``architecture'', ``algorithm''). For each method, it extracts the name, description, category (proposed, baseline, component), and key steps.

\paragraph{Experiment Extractor} Processes experiment sections to extract experimental setups, datasets used, evaluation metrics, and key results. It also identifies dataset nodes for cross-referencing.

\paragraph{Linkage Agent} Connects figures and tables to the concepts and methods they illustrate. Given a figure caption, nearby text, and a list of existing concepts, the agent determines which concepts the figure relates to and the type of relationship (illustrates, summarizes, compares, demonstrates).

The extraction proceeds in five phases: (1) concept extraction from body chunks, (2) method extraction from method sections, (3) experiment and dataset extraction, (4) figure and table linkage, and (5) inter-concept relationship discovery. Each phase updates the shared \texttt{MindGraph} data structure.

\subsection{Graph-Aware Q\&A}

The Q\&A system combines vector-based retrieval with graph traversal. The \texttt{EmbeddingStore} indexes both text chunks and node descriptions using sentence-transformers (with a simple bag-of-words fallback when unavailable). Given a question, the \texttt{GraphRetriever}:

\begin{enumerate}
    \item Retrieves the top-$k$ most similar chunks and nodes.
    \item Expands context by including 1-hop graph neighbors.
    \item Returns chunks, nodes, and connecting edges.
\end{enumerate}

The \texttt{PaperQA} agent constructs a prompt with the retrieved context, including text chunks with their section sources, relevant concept descriptions, and graph relationships. The response includes the answer, supporting sections, relevant figures and tables, and a confidence estimate.

A \texttt{locate} function allows users to find where specific items are discussed in the paper by searching across nodes, figures, tables, and text chunks, returning page numbers and context snippets.

\subsection{Coverage Verification}

To ensure nothing is silently dropped during extraction, the \texttt{CoverageChecker} produces a detailed coverage report:

\begin{itemize}
    \item \textbf{Figure coverage}: How many figures are linked to concepts or methods.
    \item \textbf{Table coverage}: How many tables are linked to results or experiments.
    \item \textbf{Section coverage}: How many sections have extracted concepts.
    \item \textbf{Equation coverage}: How many equations are linked to concepts they define.
\end{itemize}

The report includes an overall coverage score (0--100\%), lists of unlinked items with suggestions, and critical issues (e.g., ``No figures are linked to concepts/methods''). This enables quality assurance before downstream use.

\subsection{Human Verification Workflow}

The \texttt{VerificationManager} supports human-in-the-loop review:

\begin{itemize}
    \item \texttt{verify\_node}: Mark a node as human-verified.
    \item \texttt{edit\_node}: Modify node title or description.
    \item \texttt{add\_edge}: Create new relationships.
    \item \texttt{remove\_edge}: Delete incorrect relationships.
    \item \texttt{flag\_for\_review}: Flag nodes for review with a reason.
\end{itemize}

Each action is logged with timestamps, maintaining a complete edit history. Nodes carry a \texttt{verification\_status} field (auto-generated, human-verified, human-edited, or flagged) that propagates through exports.

\subsection{Export Formats}

The system exports to multiple formats for different use cases:

\begin{itemize}
    \item \textbf{JSON}: Full graph data including nodes, edges, chunks, and metadata.
    \item \textbf{Markdown}: Structured reading notes with section outlines.
    \item \textbf{Mermaid}: Mind maps and flowcharts for visualization.
    \item \textbf{HTML}: Interactive D3.js-based graph visualization.
\end{itemize}

All exports preserve traceability metadata, enabling users to navigate from any extracted element back to the original source.

\section{Implementation and Deployment}
The backend is implemented in Python with FastAPI for service endpoints and relies on standard scientific libraries for retrieval and scoring (scikit-learn, NumPy, pandas). The multi-agent pipeline is defined in \\texttt{backend/agents/discovery/pca.py}, while the refactored deterministic pipeline is implemented in \\texttt{backend/core/paperfinder.py}. Both pipelines expose functionality through API servers, including a fast discovery variant designed for low-latency responses.

The frontend is built with React and TypeScript and integrates discovery results through the API. Supabase provides authentication and persistent data storage for user profiles, communities, sessions, and paper metadata. Containerization support is provided via a Dockerfile, and deployment configurations are included for common platforms (Railway, Render, and Vercel). Environment variables control API URLs and database credentials, enabling local development or hosted deployment without code changes.

\end{document}